\documentclass[10pt,twocolumn,letterpaper]{article}

\usepackage[pagenumbers]{cvpr} 

\usepackage{graphicx}
\usepackage{amsmath}
\usepackage{amssymb}
\usepackage{booktabs}
\usepackage{mathtools}
\usepackage{comment}
\usepackage{subcaption}
\usepackage{enumitem}
\usepackage{color}
\usepackage{colortbl}
\usepackage[dvipsnames]{xcolor}
\usepackage{multirow}
\usepackage[pagebackref=true,breaklinks=true,letterpaper=true,colorlinks,bookmarks=false]{hyperref}

\usepackage{inconsolata}

\hypersetup{
   urlcolor=blue,
   citecolor=ForestGreen,
}

\usepackage[capitalize]{cleveref}
\crefname{section}{Sec.}{Secs.}
\Crefname{section}{Section}{Sections}
\Crefname{table}{Table}{Tables}
\crefname{table}{Tab.}{Tabs.}

\begin{document}

\title{EDICT: Exact Diffusion Inversion via Coupled Transformations}

\author{Bram Wallace\\
\and
Akash Gokul \\ 
Salesforce Research\\
{\tt\small \{b.wallace,agokul,nnaik\}@salesforce.com}
\and
Nikhil Naik
}
\maketitle

\begin{abstract}
Finding an initial noise vector that produces an input image when fed into the diffusion process (known as inversion) is an important problem in denoising diffusion models (DDMs), with applications for real image editing.  
The state-of-the-art approach for real image editing with inversion uses denoising diffusion implicit models (DDIMs~\cite{ddim}) to deterministically noise the image to the intermediate state along the path that the denoising would follow given the original conditioning. 
However, DDIM inversion for real images is unstable as it relies on local linearization assumptions, which result in the propagation of errors, leading to incorrect image reconstruction and loss of content. 
To alleviate these problems, we propose Exact Diffusion Inversion via Coupled Transformations (EDICT), an inversion method that draws inspiration from affine coupling layers. 
EDICT enables mathematically exact inversion of real and model-generated images by maintaining two coupled noise vectors which are used to invert each other in an alternating fashion.  
Using Stable Diffusion~\cite{ldm}, a state-of-the-art latent diffusion model, we demonstrate that EDICT successfully reconstructs real images with high fidelity. 
On complex image datasets like MS-COCO, EDICT reconstruction significantly outperforms DDIM, improving the mean square error of reconstruction by a factor of two.  
Using noise vectors inverted from real images, EDICT enables a wide range of image edits---from local and global semantic edits to image stylization---while maintaining fidelity to the original image structure. EDICT requires no model training/finetuning, prompt tuning, or extra data and can be combined with any pretrained DDM. 
Code is available at \url{https://github.com/salesforce/EDICT}.
\end{abstract}

\section{Introduction}
\label{sec:intro}

\begin{figure*}[t]
    \centering
    \includegraphics[width=\linewidth]{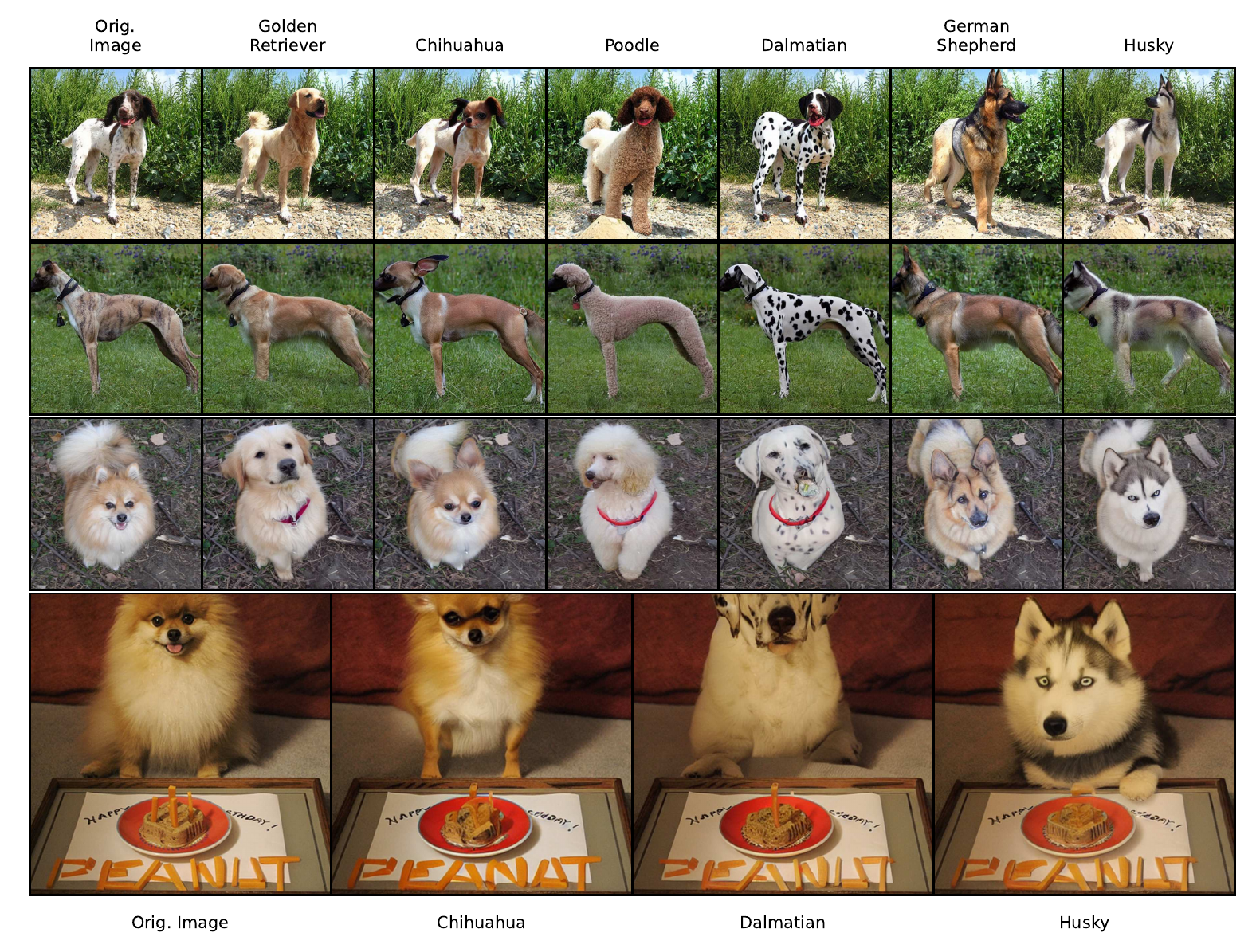}
    \caption{
    EDICT enables complex real image edits, such as editing dog breeds.
    We highlight the fine-grain text preservation in the bottom row of examples, with the message remaining even as the dog undergoes dramatic transformations.
    More examples, including baseline comparisons for all image-breed pairs, are included in the Supplementary.
    All original images from the ImageNet 2012 validation set.
    }
    \label{fig:edict_dog_editing}
\end{figure*}

Using the iterative denoising diffusion principle, denoising diffusion models (DDMs) trained with web-scale data can generate highly realistic images conditioned on input text, layouts, and scene graphs~\cite{imagen,ldm,dalle2}. 
After image generation, the next important application of DDMs being explored by the research community is that of image editing. 
Models such as DALL-E-2~\cite{dalle2} and Stable Diffusion~\cite{ldm} can perform inpainting, allowing users to edit images  through manual annotation. Methods such as SDEdit~\cite{sdedit} have demonstrated that both synthetic and real images can be edited using stroke or composite guidance via DDMs. However, the goal of a holistic image editing tool that can edit any real/artificial image using purely text has still not been achieved. 

The generative process of DDMs starts with an initial noise vector ($x_T$) and performs iterative denoising (typically with a guidance signal e.g. in the form of text-conditional denoising), ending with a realistic image sample ($x_0$). To solve the image editing problem, running the reverse of this generative process is necessary. 
Formally, this problem is known as ``inversion'' i.e., finding the initial noise vector that produces the input image when passed through the diffusion process. 

A na\"ive approach for inversion is to add Gaussian noise to the input image and perform a predefined number of diffusion steps, which typically results in significant distortions~\cite{p2p}. A more robust method is adapting Denoising Diffusion Implicit Models (DDIMs)~\cite{ddim}. Unlike the commonly used Denoising Diffusion Probabilistic Models (DDPMs)~\cite{ddpm}, the generative process in DDIMs is defined in a non-Markovian manner, which results in a deterministic denoising process.
DDIM can also be used for inversion, deterministically noising an image to obtain the initial noise vector ($x_0\rightarrow x_T$). 

\begin{figure}
    \centering
  \includegraphics[width=\linewidth]{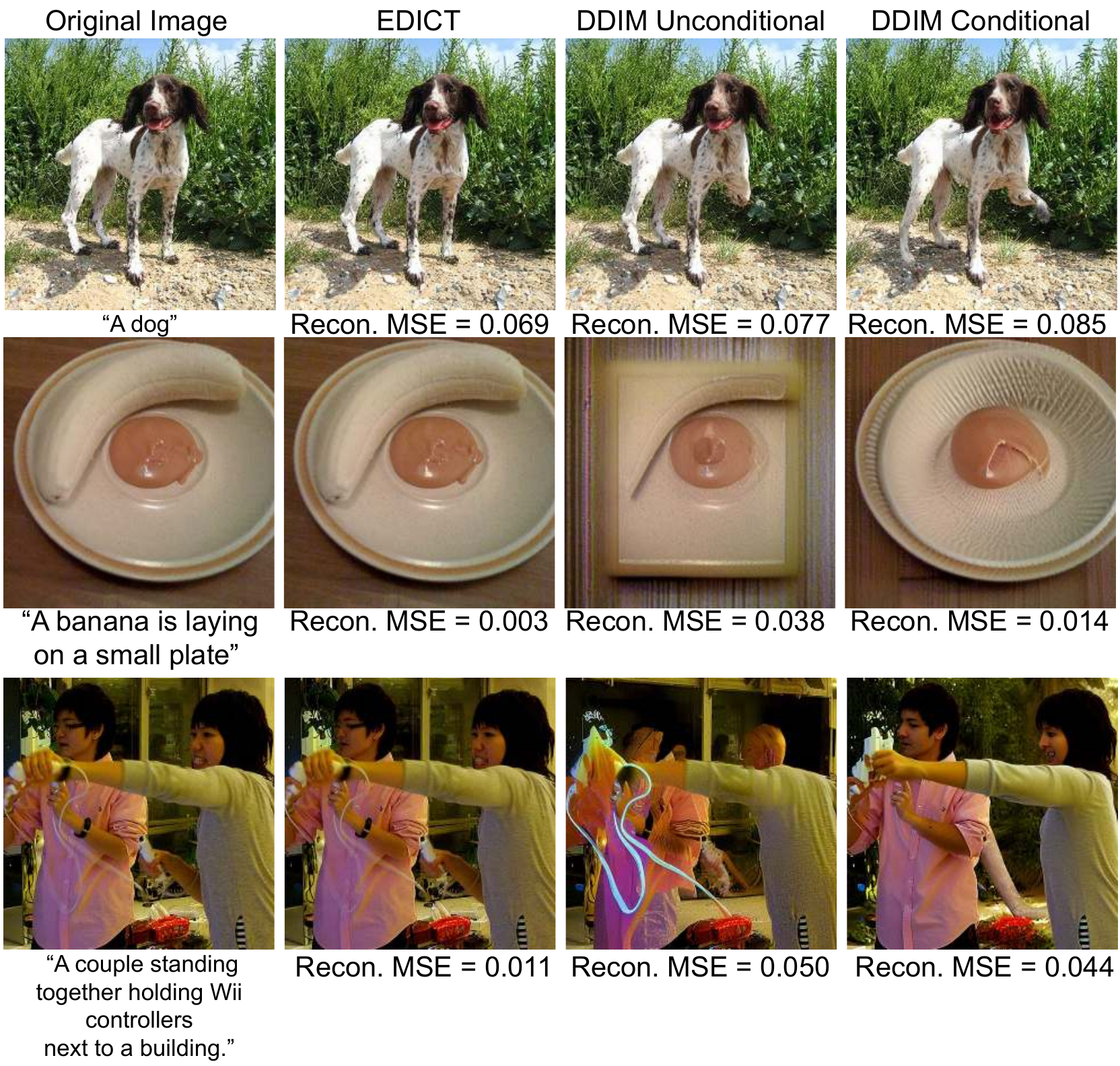}
    \caption{While both unconditional and conditional DDIM~\cite{ddim} often fail to accurately reconstruct real images, leading to loss of global image structure and/or finer details, EDICT is able to almost perfectly reconstruct even complex scenes, as seen in examples from ImageNet and COCO. All reconstructions use 50 steps. Captions used only in DDIM Conditional reconstruction with a guidance scale of 3.}
    \label{fig:ddim_failure}
\end{figure}

DDIM inversion has been used for editing real images through text methods such as DDIBs~\cite{bridges} and Prompt-to-Prompt (P2P) image editing~\cite{p2p}. After DDIM inversion, P2P edits the original image by running the generative process from the noise vector and injecting conditioning information from a new text prompt through the cross-attention layers in the diffusion model, thus generating an edited image that maintains faithfulness to the original content while incorporating the edit. However, as noted in the original P2P work~\cite{p2p}, the DDIM inversion is unstable in many cases---encoding from $x_0$ to $x_T$ and back often results in inexact reconstructions of the original image as in \cref{fig:ddim_failure}. 
These distortions limit the ability to perform significant manipulations through text as increase in the corruption is correlated with the strength of the conditioning.

To improve the inversion ability of DDMs and enable robust real image editing, we diagnose the problems in DDIM inversion, and offer a solution: Exact Diffusion Inversion via Coupled Transformations (EDICT).
EDICT is a re-formulation of the DDIM process inspired by coupling layers in normalizing flow models~\cite{glow,nice,realnvp}  that allows for mathematically {exact} inversion. By maintaining two coupled noise vectors in the diffusion process, EDICT enables recovery of the original noise vector in the case of model-generated images; and for real imagery, initial noise vectors that are guaranteed to map to the original image when the EDICT generative process is run. While EDICT doubles the computation time of the diffusion process, it can be combined with any pretrained DDM model and does not require any computationally-expensive model finetuning, prompt tuning, or multiple images. 

For the standard generative process, EDICT approximates DDIM well, resulting in nearly identical generations given equal initial conditions
For real images, EDICT can recover a noise vector which yields an exact reconstruction when used as input to the generative process. 
Experiments with the COCO dataset~\cite{lin2014microsoft} show that EDICT can recover complex image features such as detailed textures, thin objects, subtle reflections, faces, and text, while DDIM fails to do so consistently. 
Finally, using the initial noise vectors derived from a real image with EDICT, we can sample from a DDM and perform complex edits or transformations to real images using textual guidance. We show editing capabilities including local and global modifications of objects and background  and object transformations (\cref{fig:edict_dog_editing}).

\section{Related Work}
\noindent\textbf{Diffusion Models and Normalizing Flows:} Denoising diffusion models (DDMs), drawing on nonequilibrium thermodynamics~\cite{thermo}, have emerged at the forefront of image generation. Models such as GLIDE\cite{glide}, DALLE-2\cite{dalle2}, Imagen (Video)\cite{imagen,imagenvideo}, Latent/Stable Diffusion\cite{ldm}, and eDiffi\cite{ediffi} all utilize concepts borrowed from thermodynamics to hallucinate an image from pure noise by training on intermediately noised images. 
While a commonly used sampling process in DDMs is the stochastic Denoising Diffusion Probabilistic Models (DDPMs) method~\cite{ddpm}, a deterministic sampling method was introduced in Denoising Diffusion Implicit Models (DDIM)~\cite{ddim}.
Multi-step or higher-order methods that parallel DDIM have also been proposed~\cite{pndm,dpmsolver}, including methods that aim to reducing the computational time for generation~\cite{rectifiedflow,fastdpm}. 
Another class of generative models relevant to our work are normalizing flow models~\cite{nice,realnvp,glow}. In these models, an invertible mapping is learned between a latent gaussian distribution and image space. The methods of invertibility, specifically coupling layers, are used as an inspiration for our method. Invertible neural networks have been studied in areas outside of normalizing flows as well.
Neural ODEs~\cite{neuralode} have many parallels with the diffusion process and can be inverted using a variety of ODE solvers.

\noindent\textbf{Editing in Diffusion Models:}  The seminal work in applying DDMs to image editing is SDEdit\cite{sdedit} where coarse  layouts are used to guide the generative process by noising the layout to resemble an intermediately noised image.
Prompt-to-Prompt~\cite{p2p} combines query-key pairs from one prompt with values from another in the attention layers of a DDM to enable prompt-guided image editing from intermediate latents obtained by sampler inversion\cite{ddim,betterimg2img}. DiffEdit~\cite{couairon2022diffedit} edits real/synthetic images using automatically generated masks for regions of an input image that should be edited given a text query.  Kwon et al.~\cite{kwon2022diffusion} introduce style and structure losses to guide the sampling process to enable text-guided image translation. CycleDiffusion~\cite{wu2022unifying} uses a deterministic DPM encoder to enable zero-shot image-to-image translation. Another set of methods~\cite{imagic,dreambooth,textualinversion} finetune the model with the target image and/or learn a new conditioning prompt to enable indirect image editing via sampling. EDICT, our proposed approach, does not require any specialized model training/finetuning or losses, and can be paired with any pretrained DDM.

\section{Background}\label{sec:background}
\subsection{Denoising Diffusion Models}\label{sec:bg_diffusion}

DDMs are trained on a simple denoising objective.
A set of timesteps index a monotonic strictly increasing noising schedule $\{\alpha_t\}^{T}_{t=0}, \alpha_T=0, \alpha_0=1$. 
Images (or autoencoded latents) $x \in X$ are noised with draws $\epsilon \sim N(0,1)$ according to the noising schedule following the formula
\begin{equation}
    x_t = \sqrt{\alpha_t} x + \sqrt{1 - \alpha_t} \epsilon
\end{equation}
The time-aware DDM $\Theta$ is trained on the objective $MSE(\Theta(x_t, t, C), \epsilon)$ to predict the noise added to the original image where $C$ is a conditioning signal (typically in the form of a text embedding) with some degree of dropout to the null conditioning $\emptyset$.
To generate a novel image from a gaussian draw $\epsilon_T \sim N(0,1)$, partial denoising is applied at each $t$.
The most common sampling scheme is that of DDIM~\cite{ddim} where intermediate steps are calculated as
\begin{align}\label{eq:ddim}
    \begin{split}
    x_{t-1} =& \sqrt{\alpha_{t-1}} \frac{x_t - \sqrt{1 - \alpha_t} \Theta(x_t, t, C)}{\sqrt{\alpha_t}} \\
    &+ \sqrt{1 - \alpha_{t-1}} \Theta(x_t, t, C)
    \end{split}
\end{align}
In practice, for text-to-image models to hallucinate from random noise an $x_0$ that matches conditioning $C$ to desired levels, the model has to be biased more heavily towards generations aligned with $C$.
To do so, a pseudo-gradient $G \cdot (\Theta(x_t, t, C) - \Theta(x_t, t, \emptyset))$ is added to the  unconditional prediction $\Theta(x_t, t, \emptyset)$ to up-weight the effect of conditioning, where  $G$ is a weighting parameter, 
Substituting $\Phi(x_t, t, C, G)=\Theta(x_t, t, \emptyset) + G \cdot (\Theta(x_t, t, C) - \Theta(x_t, t, \emptyset))$ into the prior equation for the $\Theta$ term, we simplify the notation $\Phi(x_t, t, C, G) \xrightarrow{} \epsilon(x_t, t)$ and rewrite the previous equation as  $x_{t-1} = a_t x_t + b_t \epsilon(x_t, t)$ where
\begin{align}
    & a_t = \sqrt{\alpha_{t-1}/{\alpha_t}} \\
    & b_t = -\sqrt{ {\alpha_{t-1}(1-\alpha_t)}/{\alpha_t}} + \sqrt{1 - \alpha_{t-1}}
\end{align}

\subsection{Denoising Diffusion Implicit Model (DDIM)}\label{sec:ddim_inversion}
As noted in DDIM~\cite{ddim}, the above denoising process is approximately invertible; that is $x_t$ is approximately recoverable from $x_{t-1}$
\begin{align}
    x_t  = \frac{x_{t-1} - b_t \epsilon(x_t, t)}{a_t}
    \approx \frac{x_{t-1} - b_t \epsilon(x_{t-1}, t)}{a_t}
\end{align}
where the approximation is a linearization assumption that $\epsilon(x_t, t) \approx \epsilon(x_{t-1}, t)$ (necessary due to the discrete nature of both computation and the underlying noise schedule).
This corresponds with reversing the Euler integration which is a first-order ODE solver.
More sophisticated solvers such as multi-step Euler \cite{pndm} have been shown to stabilize the generative process, and correspondingly the deterministic inversion process,  with fewer time steps.
However, such methods are also approximations where the inversion accuracy ultimately relies on the strength of the linearization assumption and the reconstruction is not exactly equal.
This assumption is largely accurate for unconditional DDIM models, but the pseudo-gradient of classifier-free guidance $G \cdot (\Theta(x_t, t, C) - \Theta(x_t, t, \emptyset))$ is inconsistent across time steps as shown in the Supplementary. 

While unconditional reconstructions have relatively insignificant errors (\cref{fig:ddim_failure}), \textit{conditional} reconstructions are extremely distorted when noised to high levels.
This phenomenon was noted in \cite{p2p}, where the guidance scale must be heavily downweighted in order for inversions on real-world images to be stable, thus limiting the strength of edits.
Obtaining an $x_t$ from $x_0$ allows for the generative process to be run with novel conditioning. 
In SDEdit~\cite{sdedit}, this process is done stochastically to obtain broad sample diversity, at the cost of controllability and faithfulness to the original image contents.
In contrast, the inverse DDIM process produces a unique $x_t$ from a single $x_0$ in a deterministic manner, yielding only one sample but enabling higher strength edits while preserving finer-grain structure and content.

\subsection{Affine Coupling Layers}\label{sec:bg_coupling}
Affine Coupling Layers (ACL) are invertible neural network layers introduced in \cite{nice,realnvp} and used in other normalizing flow models such as Glow~\cite{glow}. 
The layer input $z$, is split into two equal-dimensional halves $z_a$ and $z_b$.
A modified version of $z_a$ is then calculated, according to:
\begin{align}\label{eq:background_coupling}
    z'_a = \Psi(z_b) z_a + \psi(z_b)
\end{align}
where $\Psi$ and $\psi$ are neural networks.
The layer output $z'$ is the concatenation of $z'_a$ and $z_b$ in accordance with the original splitting function.
ACL can parameterize  complex functions and $z$ can be {exactly} recovered given $z'$:
\begin{align}
    z_a = ({z'_a - \psi(z_b)})/{\Psi(z_b)}
    \label{eq:invert}
\end{align}
 Noting the similarity of equations~\ref{eq:background_coupling} and~\ref{eq:invert} to the simplified form of \cref{eq:ddim}, we parallel this construction in our method (described next) where two separate quantities are tracked and alternately modified by transformations that are affine with respect to the original modified quantity and a non-linear transformation of its counterpart.

\begin{figure}[t]
  \centering
  \includegraphics[width=0.75\linewidth]{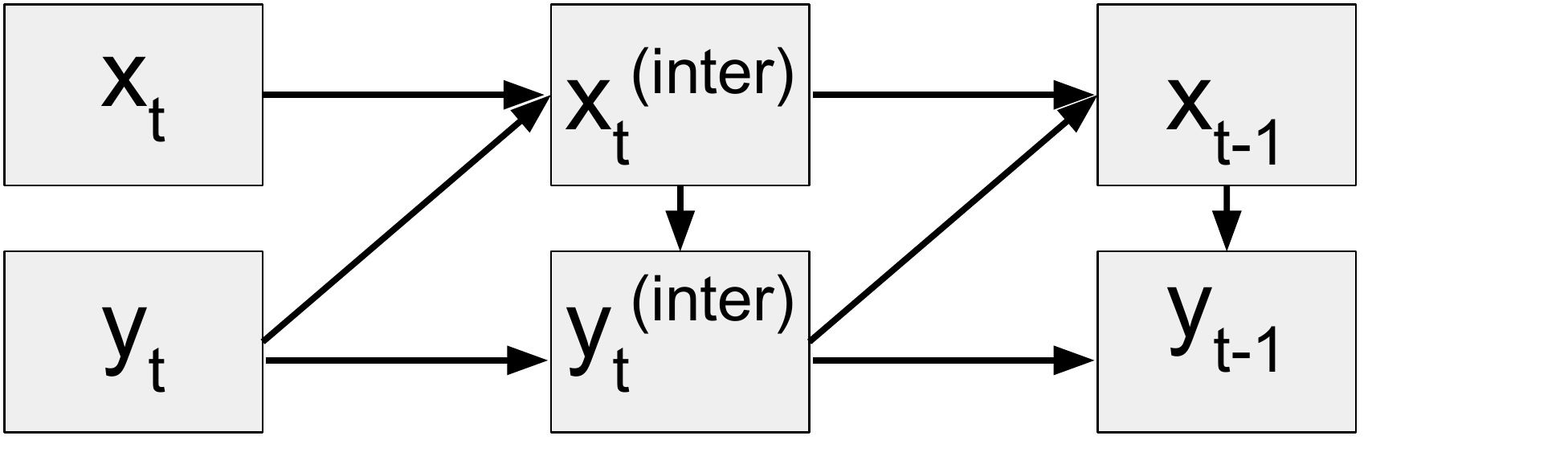}
   \caption{
   Information flow of EDICT.
   Denoising process (sampling) follows the arrows forwards, sampler inversion backwards.
   From $(x_t,y_t)$, $(x_{t-1}, y_{t-1})$ can be calculated while holding a single member of each sequence in memory at a time.
   All steps are invertible, so $(x_t,y_t)$ can be exactly recovered from $(x_{t-1}, y_{t-1})$, as opposed to current methods which compute an approximation.
   }
   \label{fig:coupling}
\end{figure}
\section{Exact Diffusion Inversion via Coupled Transformations (EDICT)}
\subsection{Making an Invertible Diffusion Process}
As summarized in \cref{sec:bg_coupling}, affine coupling layers  track {two} quantities which can then be used to invert each other. These two quantities are partitions of a latent representation with a network specifically designed to operate in a fitting alternating manner.
Without training of a new DDM, this method can not be applied out-of-the-box to the forward diffusion process.
We consider the simplified form of the forward step equation from \cref{sec:bg_diffusion} below
 \begin{equation}
     x_{t-1} \coloneqq a_t x_{t} + b_t \epsilon(x_t, t)
 \end{equation}
 If the noise prediction term, $\epsilon(x_t, t)=\varepsilon$ was independent of $x_t$, this would be an {affine} function in both $x_t$, and $\varepsilon$.
 Paralleling \cref{eq:background_coupling}, by creating a new variable $y_t=x_t$ the stepping equation fits the desired form.
 Consider performing this computation, so we have the variables $x_t,\ \ y_t=x_t,\ \ x_{t-1} = a_t x_{t} + b_t \epsilon(y_t, t)$.
 $x_t$ can be recovered exactly from $x_{t-1}$ in the non-trivial form: 
 \begin{equation}
     x_t = ({x_{t-1} - b_t \cdot \epsilon(y_t, t) })/{a_t}
 \end{equation}
 
 \noindent$y_t=x_t$ is trivial and will not be true in the general case.
Now consider the initialization of the reverse (denoising) diffusion process, where $x_T \sim \mathcal{N}(0,1)$, we similarly initialize $y_T=x_T$.
 Following the above process, we define the update rule
 \vspace{-5pt}
 \begin{equation}
 \begin{split}
     x_{t-1} = a_t x_{t} + b_t \cdot \epsilon(y_t, t) \\
    y_{t-1} = a_t y_{t} + b_t \cdot \epsilon(x_{t-1}, t)
    \end{split}
 \end{equation}
Note that the noise prediction term in the second line is a function of the \textit{other} sequence value at the {next} timestep.
Only one member of each sequence ($x_i,y_j$) must be held in memory at any given time.
The sequences can be recovered exactly according to
\vspace{-5pt}
\begin{equation}
 \begin{split}
     y_{t} = {(y_{t-1} - b_t \cdot \epsilon(x_{t-1}, t))}/{a_t} \\
     x_{t} = {(x_{t-1} - b_t \cdot \epsilon(y_t, t))}/{a_t} 
    \end{split}
 \end{equation}
As illustrated in \cref{fig:coupling}, the entire sequence can be reconstructed from any two adjacent $x_i$ and $y_i$.
 In sum, our method re-uses the linearization assumption of Euler DDIM inversion, that $\epsilon(x_t, t) \approx \epsilon(x_{t-1}, t)$, but crucially does not rely on it for invertibility, guaranteeing recovery up to machine precision.
We note that the functional form of our approach bears similarities to the ODE solver Heun's method where derivative values at initial predictions are used to refine predictions, but due to the need for invertibility we cannot exploit this for more accurate/faster sampling.

\begin{figure}[t]
    \centering
    \includegraphics[width=1.0\columnwidth]{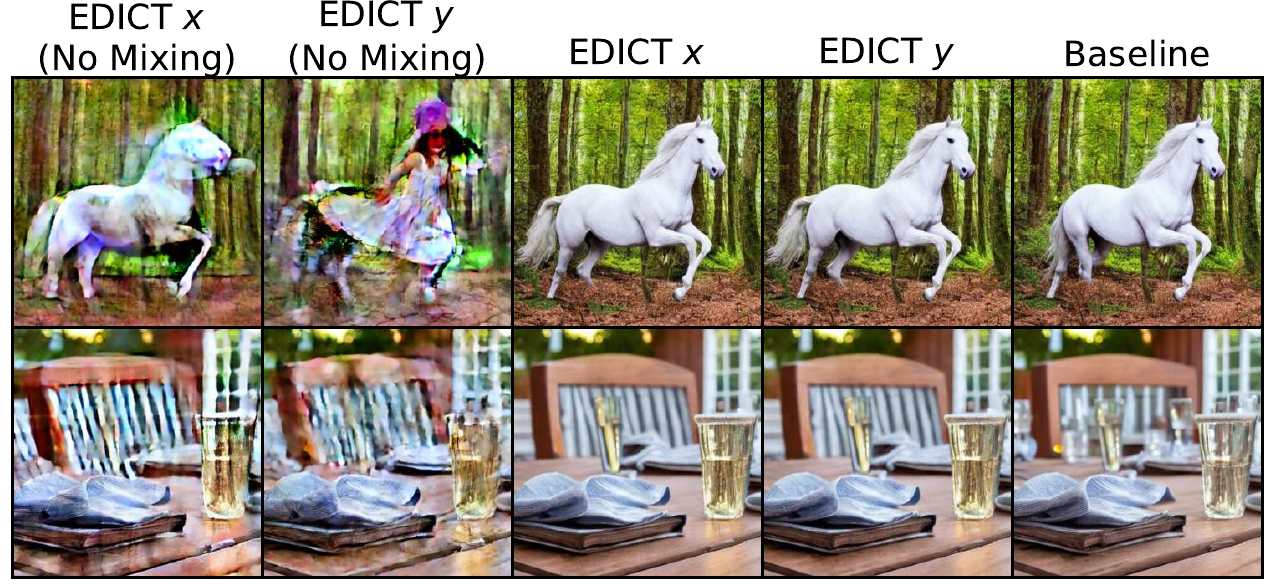}
    \caption{
    Images generated from the same text prompt and random seed.
    \textbf{EDICT $x/y$ (No mixing)} employs coupling layers but not intermediate mixing layers, resulting in distorted and inconsistent images as $x$ and $y$ diverge.
    \textbf{EDICT} is our full method with the sequences explicitly contracted together, resulting in  identical images that match DDIM (\textbf{Baseline}) in quality and  composition.
    Top row prompt: \textit{A white horse galloping through a forest}
    Bottom row prompt: \textit{A couple of glasses are sitting on a table} 
    }
    \label{fig:edict_gens}
\end{figure}

\subsection{Stabilization}
\begin{figure}[t]
  \centering
  \includegraphics[width=\linewidth]{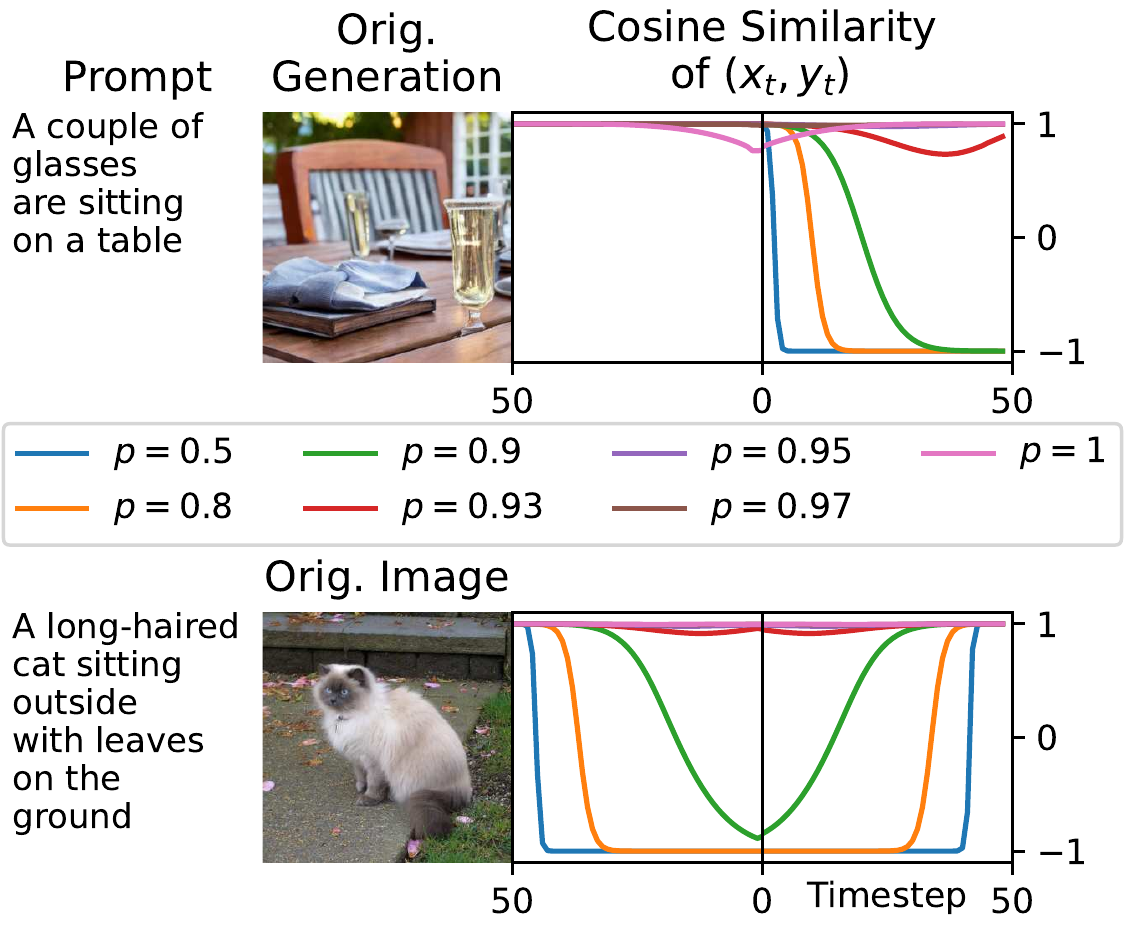}

   \caption{
   We visualize the effect of the mixing coefficient $p$.
   Top row: an EDICT-generated image with $p=0.93$, for each listed $p$ a generation is computed from $T=50$ to $T=0$ followed by reconstructing initial latents ($T=50$).
   Cosine similarity between $(x_t, y_t)$ is computed at each step.
   $p=0.97$ suffices for generative convergence, but the \textit{inverse} process diverges when $p$ is too small.
   Repeated dilation can exponentially exaggerate small floating-point differences, annulling the theoretical guarantee.
   These trends hold in the bottom row for a non-generated image put through the EDICT inversion/generation process.
   While all reconstructions ultimately are aligned, a sufficiently large $p$ is needed for the latents to maintain alignment throughout.
   }
   \label{fig:divergence}
\end{figure}

While our method assures invertibility by design, realism and faithfulness to the original diffusion process are not automatic. When naively applied for a typical, low number of DDIM steps (e.g., $T = 50$), the sequences $x_t$ and $y_t$ can diverge (\cref{fig:edict_gens}).
This is a result of the strong linearization assumption not holding in practice.
To alleviate this problem, we introduce intermediate \textit{mixing layers} after each diffusion step computing weighted averages of the form 
\begin{equation}
x' = px + (1-p)y,\ 0\le p \le 1
\end{equation}
which are invertible affine transformations.
Note that this averaging layer becomes a \textit{dilating} layer during deterministic noising; the inversion being 
\vspace{-5pt}
\begin{equation}
x = \frac{x' - (1-p)y}{p} 
\end{equation}
A high (near 1) value of $p$ results in the averaging layer not being strong enough to prevent divergence of the $x$ and $y$ series during denoising, while a low value of $p$  results in a numerically unstable exponential dilation in the backwards pass (\cref{fig:divergence}). Note that in both cases that our generative process remains {exactly} mathematically invertible with no further assumptions, but there is a degradation in utility of the results.  Typically we employ $p=0.93$, with values in the interval $[0.9, 0.97]$ generally being effective for 50 steps.

\subsection{Complete Summary of the Method}
We dub our presented process EDICT: \textbf{E}xact \textbf{D}iffusion \textbf{I}nversion via \textbf{C}oupled \textbf{T}ransformations.
In sum, EDICT uses a combination of ``coupling'' and ``averaging/dilating'' steps for exact inversion of the diffusion process. 
Given $x_t$ and $y_t$, we  calculate the denoising process by
\begin{align}
\begin{split}\label{eq:edict_forward}
    x^{inter}_t &=  a_t \cdot x_{t} + b_t \cdot  \epsilon(y_t, t) \\
    y^{inter}_t &=  a_t \cdot  y_{t} + b_t  \cdot  \epsilon(x^{inter}_t, t) \\
    x_{t-1} &= p \cdot x^{inter}_t + (1-p) \cdot y^{inter}_t \\
    y_{t-1} &= p \cdot y^{inter}_t + (1-p) \cdot x_{t-1} \\
\end{split}
\end{align}
and the deterministic noising inversion process by:
\begin{align}
\begin{split}\label{eq:edict_reverse}
    y^{inter}_{t+1} &= ({y_{t} - (1-p) \cdot x_{t}})/{p} \\
    x^{inter}_{t+1} &= ({x_{t} - (1-p) \cdot y^{inter}_{t+1}})/{p} \\
    y_{t+1} &= ({y^{inter}_{t+1} - b_{t+1}\cdot  \epsilon(x^{inter}_{t+1}, t+1)})/{a_{t+1}} \\
    x_{t+1} &= ({x^{inter}_{t+1} - b_{t+1} \cdot \epsilon(y_{t+1}, t+1)})/{a_{t+1}} \\
\end{split}
\end{align}
Recall that the conditioning $C$ is implicitly included in the $\epsilon$ terms.
In practice, we alternate the order in which the $x$ and $y$ series are calculated at each step in order to symmetrize the process with respect to both sequences.
We cast all operations to double floating point precision (from the native half precision) to mitigate roundoff floating point errors.

\subsection{Image Editing}
Given an image $I$, we edit the semantic contents to match text conditioning $C_{target}$.
We describe the current content by text $C_{base}$ in a parallel manner to $C_{target}$.
We compute an autoencoder latent $x_0 = VAE_{enc}(I)$, initializing $y_0=x_0$.
We run the deterministic noising process of \cref{eq:edict_reverse} on $(x_0, y_0)$ using text conditioning $C_{base}$ for $s \cdot S$ steps, where $S$ is the number of global timesteps and $s$ is a chosen editing strength.
This yields partially ``noised'' latents $(x_t, y_t)$ which are not necessarily equal and, in practice, tend to diverge by a small amount due to linearization error and the dilation of the mixing layers.
These intermediate representations are then used as input to \cref{eq:edict_forward} using text condition $C_{target}$, and identical step parameters, $(s, S)$.
The resulting image outputs $(VAE_{dec}(x^{edit}_0), VAE_{dec}(y^{edit}_0))$ are empirically nearly identical as seen in \cref{fig:edict_gens}, this is by design of the method and in particular, the mixing layers.
For all methods we find that a guidance scale of 3 performs well (as opposed to the standard 7.5 for generation).

\begin{figure*}[t]
    \centering
    \includegraphics[width=\linewidth]{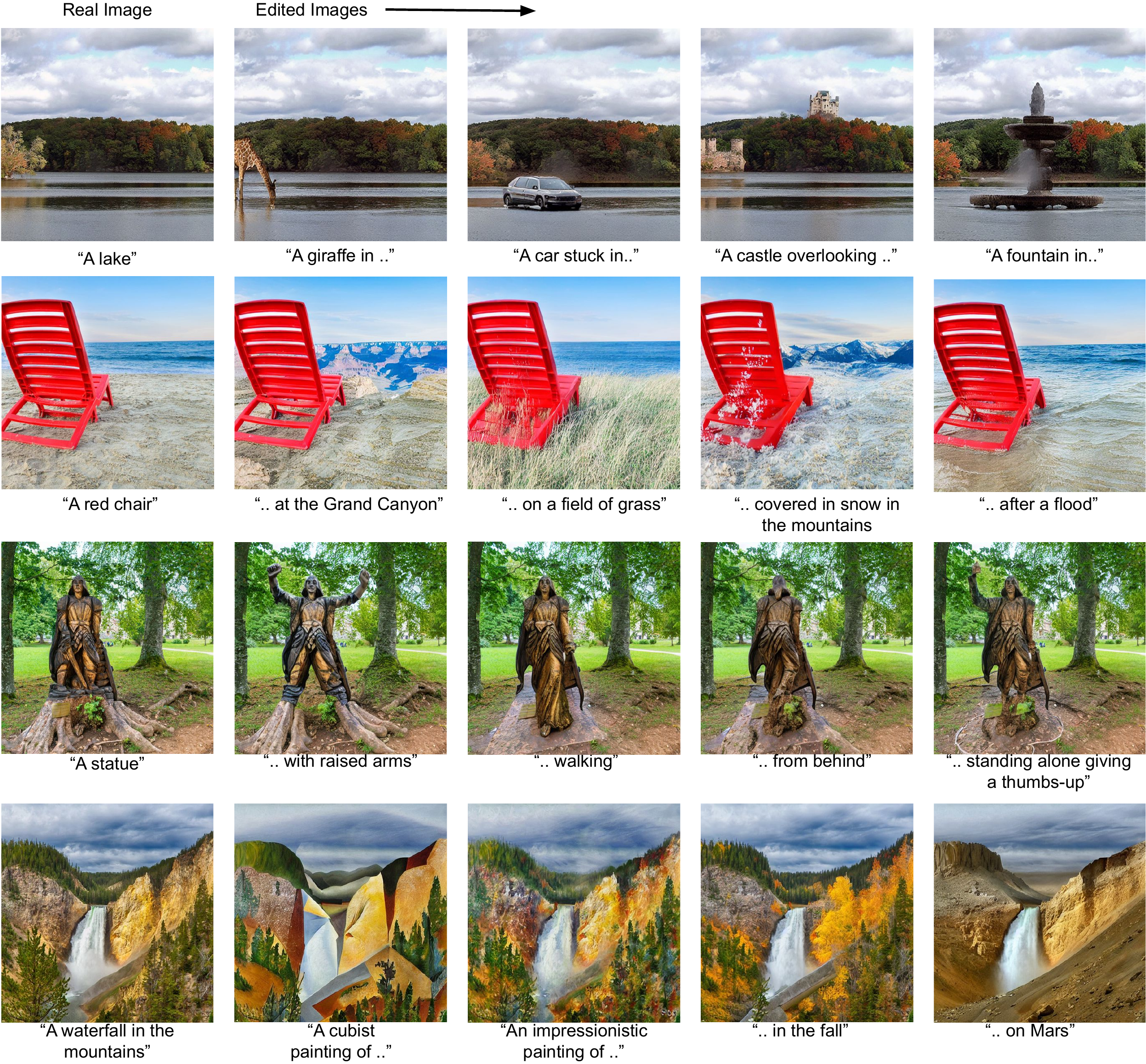}
    \caption{Examples of the strength and varied editing ability of EDICT.
    From top to bottom: object additions, global context changes with object preservation, deformations while preserving identity, and global style changes.
    The bidirectional reversibility of EDICT's process allows for large-scale changes to the image while maintaining auxiliary details.
    }
    \label{fig:edict_editing}
\end{figure*}

  \begin{table}
  \centering
  \resizebox{1\linewidth}{!}{
  \begin{tabular}{c|c c c c c}
  \multicolumn{6}{c}{COCO Reconstruction Error (MSE)} \\
    \toprule
    \multirow{2}{*}{Method} & \multirow{2}{*}{LDM AE} & EDICT & EDICT & DDIM & DDIM \\
    & & (UC) & (C) & (UC) & (C) \\
    \midrule
    50 Steps & 0.015 & 0.015 & 0.015 & 0.030 & 0.420 \\
    200 Steps & 0.015 & 0.015 & 0.015 & 0.023 & 0.497 \\
    \bottomrule
  \end{tabular} 
  }
  \caption{
  Mean-square error reconstruction results for the COCO validation set using the first listed prompt as conditioning with full-strength guidance. The latent diffusion model autoencoder (LDM AE), which is the autoencoder used to compute latents for reconstruction is the lower bound on reconstruction error.
  Values for more steps are provided in the Supplementary.
  }
  \label{tab:recon}
\end{table}

 \section{Experiments}
We now describe results using the Stable Diffusion 1.4 latent diffusion model~\cite{ldm}. 
\subsection{Image Reconstruction}
We demonstrate the exact invertibility of EDICT using the MS-COCO-2017 validation set ($n=5,000$), which contains both simple object-centric images and complex scene images~\cite{lin2014microsoft}. 
Given an image-caption pair, inverted latents are calculated and used to reconstruct the image.
Mean-square error is calculated on pixels normalized to $[-1,1]$ and averaged across all images in the dataset.
This process is performed both with and without the text as conditioning (C vs. UC). For COCO, we use the first listed prompt as conditioning. The LDM autoencoder reconstruction error serves as a lower bound.  
EDICT maintains complete latent recovery in all examples for both 50 and 200 steps, with error 50-75\% that of DDIM (UC) (\cref{tab:recon}). 
DDIM (C) is unstable for inversions (also noted in~\cite{p2p}), which results in an error an order of magnitude greater than any other, with failure to reconstruct as shown in \cref{fig:ddim_failure}.

 \begin{figure*}
    \centering
    \includegraphics[width=\linewidth]{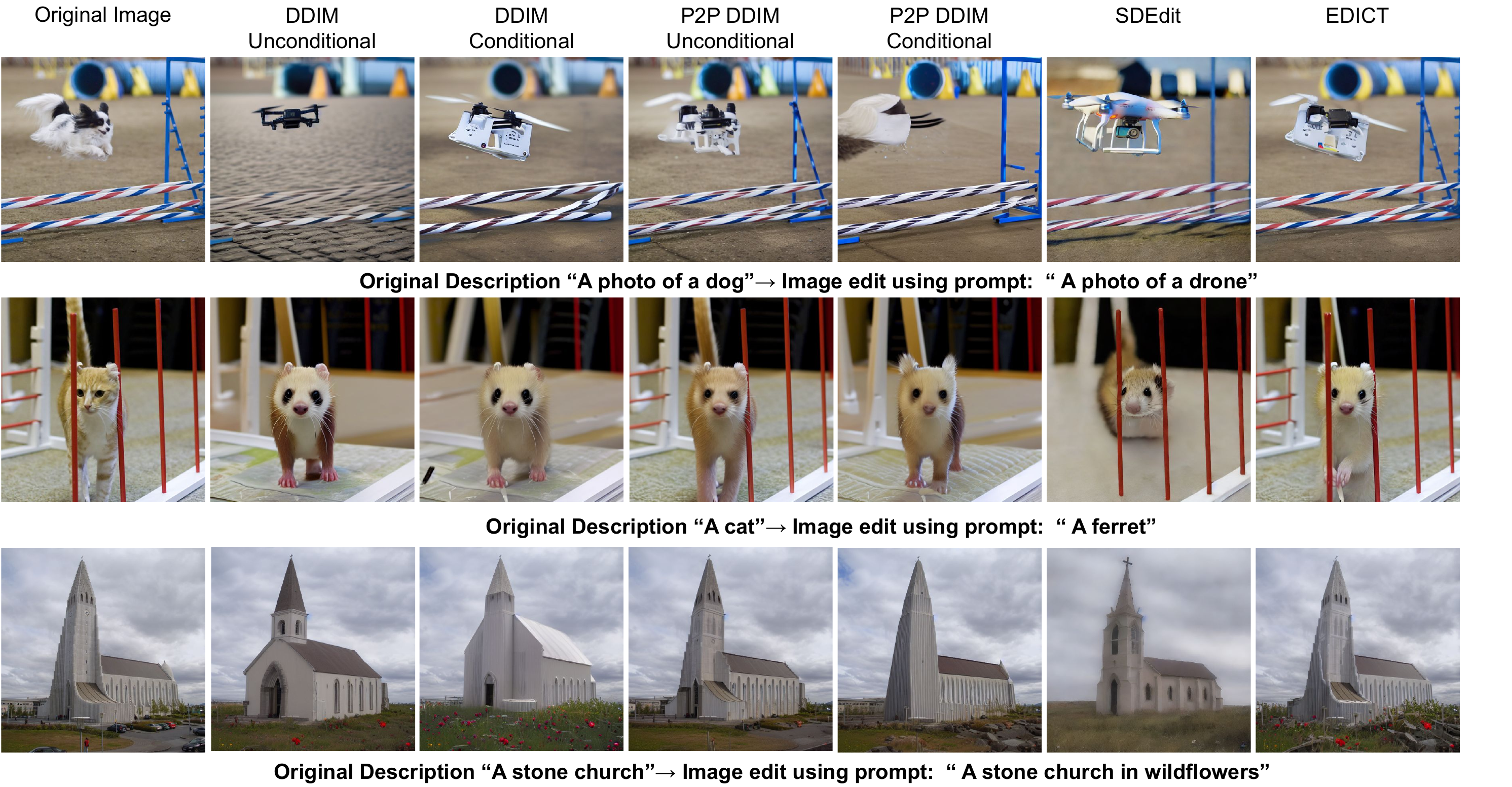}
    \caption{
    EDICT compared to baseline editing methods.
    EDICT demonstrates superior preservation of unedited components while still performing the requested semantic edit.
    Further visual comparison between EDICT and baselines is provided in the Supplementary.
    }
    \label{fig:baseline_comp}
\end{figure*}

 \subsection{Image Editing}
We show EDICT's ability to perform complex editing tasks on real images in \cref{fig:edict_editing}. 
In the first row, a diverse set of objects are added to a lake scene, demonstrating \textbf{object addition}.
In the \textit{giraffe} and \textit{car} examples, we see interaction between introduced and original objects.
Additionally, both the \textit{giraffe} and \textit{castle} examples capture reflections of the added objects, while the pattern of the water is maintained.
Throughout all edits, details such as the cloud patterns and patches of tree color are preserved.
The second row shows \textbf{object-preserving global changes}. A chair is placed into a variety of settings while keeping near-perfect identity and detail, even when occlusions are generated (\textit{grass} and \textit{snow}). 
In all examples the chair keeps a realistic footing, despite ground changes.

In the third row, we demonstrate that EDICT is able to perform \textbf{object deformations}, a challenging class of edits for DDM methods, as many broad compositional components are determined very early in the generation~\cite{p2p}. The previously most successful method for these types of edits, Imagic\cite{imagic} requires model finetuning.
EDICT makes the sculpture assume a broad set of poses, spatially changing the semantic map in a refined  way.
Novel views, such as \textit{``A statue from behind"} are able to be plausibly rendered
Across these large-scale edits, fine-grained details such as the face and dress of the statue --as well as the foliage and path-- are preserved.
In the fourth row, we show that EDICT is able to perform \textbf{global style changes} while maintaining layout and details where appropriate.
The layout is nearly identical across images (note the preserved cloud pattern), but can be changed when needed (e.g., the lack of trees in the \textit{Mars} example).
Specific art-styles are capable of being generated, including challenging concepts such as cubism.

In Fig.~\ref{fig:edict_dog_editing}, EDICT makes \textbf{semantic entity edits} to and from a variety of dog breeds.
It proves adept at holding the original subject pose, including in the third row where the dog is viewed in a very atypical position.
The realism in the \textit{Chihuahua} examples is particularly interesting, due to the relatively small size of the breed.
We again highlight the preservation of details such as the background foliage or ground in the upper three rows.
The fourth row is of specific interest as small text is preserved across examples (a typical failure case for DDIM).\\

\begin{figure}
    \centering
    \includegraphics[width=\linewidth]{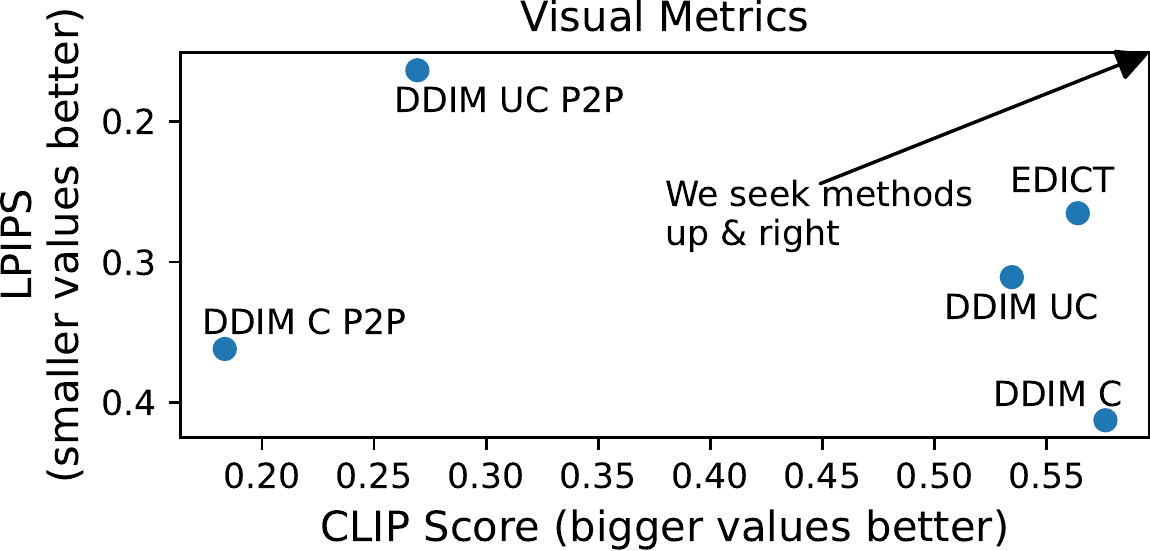}
    \caption{
   We quantitatively benchmark all methods on image editing using images from five ImageNet mammal classes (\textit{African Elephant, Ram, Egyptian Cat, Brown Bear, and Norfolk Terrier}).
   Four experiments are performed, one swapping the pictured animal's species to each of the other classes (20 species editing pairs in total), two contextual changes (\textit{A [animal] in the snow} and \textit{A [animal] in a parking lot}), and one stylistic (\textit{An impressionistic painting of a [animal]}).
   The LPIPS\cite{lpips} (visual similarity) score to the original image and a CLIP score \cite{clip} (semantic similarity) to a set of five related text queries is computed per edit.
   Metrics are averaged across images and experiments, details are provided in the Supplementary.
   EDICT largely maintains the CLIP score (0.56) of the best competing methods while substantially improving the LPIPS (0.27), quantitatively demonstrating the ability to perform state-of-the-art content changes with superior fidelity.
    }
    \label{fig:visual_metrics}
\end{figure}

\noindent\textbf{Baseline Comparison:} In \cref{fig:baseline_comp}, we demonstrate EDICT's superior performance to other DDM sampling-based methods for image editing: conditional and unconditional DDIM inversion,  prompt-to-prompt image editing, and SDEdit (as a stochastic baseline).
All methods are run with 50 steps (we do not observe improvement of baselines when the number of steps are increased, see Supplementary). Since methods that  require model finetuning or prompt tuning~\cite{imagic,dreambooth,textualinversion} can be combined with EDICT, we view them as complementary to, rather than competitive with, EDICT, and do not perform a direct comparison with them. 
We quantitatively compare visual metrics of edits in \cref{fig:visual_metrics}.

 \section{Discussion}
\noindent\textbf{Limitations and Future Work:} 
EDICT is deterministic unlike methods such as SDEdit and only outputs one generation per image-prompt pair.
Also, the computational time is approximately twice that of a baseline DDIM process.
As with all editing methods, performance can vary across inputs in a hard-to-predict manner and sometimes requires careful prompt selection. For future work, we note that the induced latent space constructed by the inversion process admits operations such latent interpolation\cite{sddream} which has not been widely applied to real images. 
As in prompt tuning\cite{prompttuning}, formalizing the process of prompt selection could further improve EDICT.
 Adding a controllable degree of randomness to EDICT could yield multiple candidate generations that still satisfy the desired properties.\\

\noindent\textbf{Ethics:} Like other image generation and editing models, EDICT will produce images that may reflect the socioeconomic biases of the training data or images that could be considered inappropriate. Image editing methods can also utilized for malicious purposes, including harassment and misinformation spread. Practitioners utilizing EDICT or its methodologies in a production setting should consider these limitations. An in-depth discussion on the ethics of image generation can be  found in Imagen~\cite{imagen}.


{\small
\bibliographystyle{ieee_fullname}
\bibliography{egbib}
}

\clearpage

\crefname{section}{Sec.}{Secs.}
\Crefname{section}{Section}{Sections}
\Crefname{table}{Table}{Tables}
\crefname{table}{Tab.}{Tabs.}
\crefname{figure}{Fig. S}{ Fig. S}

\crefformat{figure}{Figure S#2#1#3}
\crefformat{table}{Table S#2#1#3}
\renewcommand\figurename{Figure S}
\renewcommand\thefigure{\unskip\arabic{figure}}

\renewcommand\tablename{Table S}
\renewcommand\thetable{\unskip\arabic{table}}

\appendix

\noindent \textbf{The supplementary section is organized as follows:} \\
\cref{sec:sec1}: Details of the quantitative evaluation of image editing methods \\
\cref{sec:sec2}:  Additional analysis of DDIM instability. \\
\cref{sec:sec3}: Additional image editing results with EDICT \\
\cref{sec:sec4}: Additional image reconstruction results with EDICT

\section{Quantitative Experiment Details}\label{sec:sec1}

We sample from 5 ImageNet classes (\textit{African Elephant, Ram, Egyptian Cat, Brown Bear, and Norfolk Terrier}, validation set).
Four experiments are performed, one swapping the pictured animal's species to each of the other classes (20 species editing pairs in total), two contextual changes (\textit{A [animal] in the snow} and \textit{A [animal] in a parking lot}), and one stylistic (\textit{An impressionistic painting of a [animal]}).
The prompt for the species edit is simply \textit{A [animal]}.
Throughout, base prompts are of form \textit{A [animal]}.
Edits are performed with inversion strength $s=0.8$ and steps $S=50$.

For computing CLIP score in the species example, the 5 text queries are of identical form \textit{A [animal]}.
CLIP text queries for other edits are as follows:

\textit{A [animal] in the snow}
\begin{itemize}
    \item An animal in the snow
    \item An animal in the sun
    \item An animal in the rain
    \item An animal in a sand storm
    \item An animal in the ocean
\end{itemize}

\textit{A [animal] in a parking lot}
\begin{itemize}
    \item An animal in a parking lot
    \item An animal in the wild
    \item An animal in a shopping mall
    \item An animal in the ocean
    \item An animal on a football field
\end{itemize}

\textit{A impressionistic painting of a [animal]}
\begin{itemize}
    \item An impressionistic painting of an animal
    \item A photograph of an animal
    \item A crayon drawing of an animal
    \item A digital rendering of an animal
    \item A pencil drawing of an animal
\end{itemize}

\begin{figure*}
    \centering
    \includegraphics[width=0.9\linewidth]{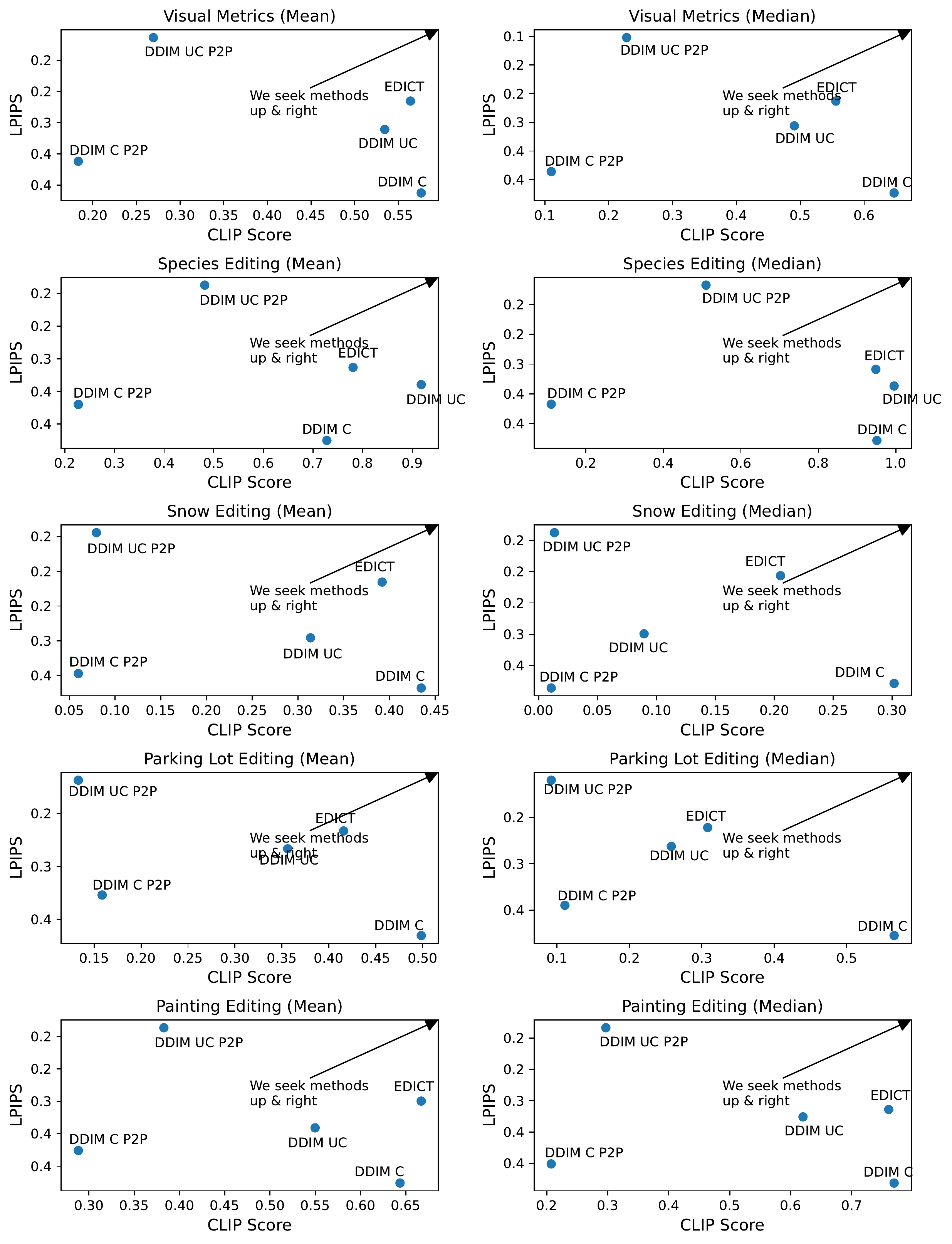}
    \caption{Median and individual plots of visual metrics.
    Pareto-optimiality is achieved in all cases and for 3 of the 4 experiments (all but species editing) EDICT improves upon DDIM UC in both metrics while far outperforming DDIM UC P2P in CLIP score (achievement of edit) and DDIM C in LPIPS (perceptual faithfullness to the original image).
    }
    \label{fig:quant}
\end{figure*}

We plot the mean and median metrics for baselines on each individual benchmark experiment as well as the mean-average and median-average across experiments in \cref{fig:quant}.

\section{Misalignment of Pseudo-Gradient}\label{sec:sec2}

\begin{figure*}
    \centering
    \includegraphics[width=0.8\linewidth]{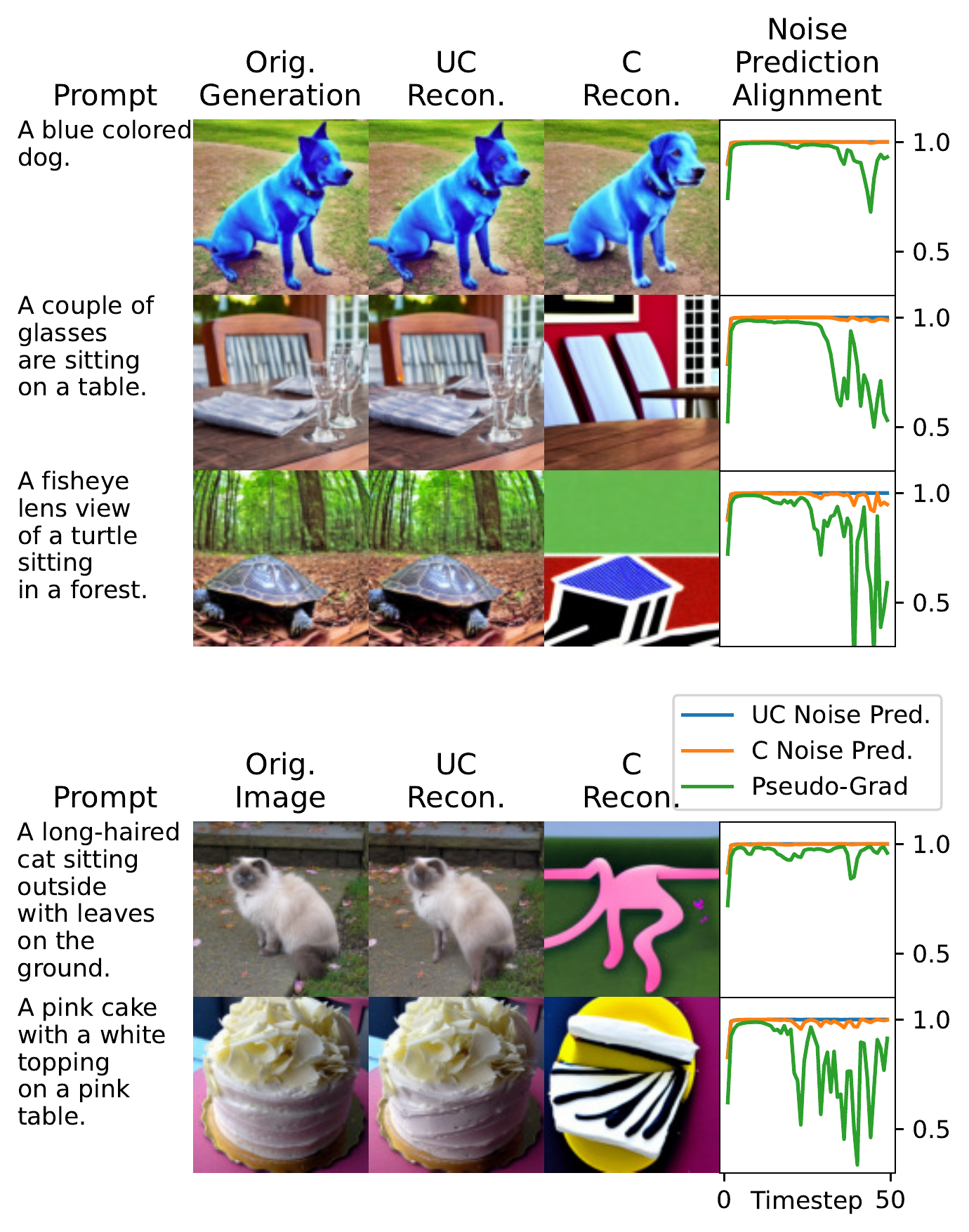}
    \caption{
    Top panel: Images are generated from a given prompt and displayed in \textit{Orig. Generation}. 
    The generation process is then inverted and re-ran both unconditionally and conditionally.
    While the unconditional reconstructions are near perfect, the conditional reconstructions suffer from varying degrees of instability.
    In the right column, we plot the cosine similarity of the noise prediction component at each timestep to the noise component at the previous timestep.
   Lower panel: the inversion-reconstruction is performed for real (non-generated) images.
    For both the unconditional and conditional component (particularly the former) we see near-perfect alignment across steps justifying the linearization assumption.
    For the pseudo-gradient term, however, at higher $t$ (further-noise timesteps) the gradient becomes extremely inconsistent across timesteps.
    This explains the lack of stability in vanilla conditional DDIM inversion, the lack of consistency in the pseudo-gradient is counter to the linearization assumption and as such reconstructions are not faithful.
    All experiments run with generation-strength guidance scale of $G=7$.
    }
    \label{fig:inversion_50}
\end{figure*}

\begin{figure*}
    \centering
    \includegraphics[width=.8\linewidth]{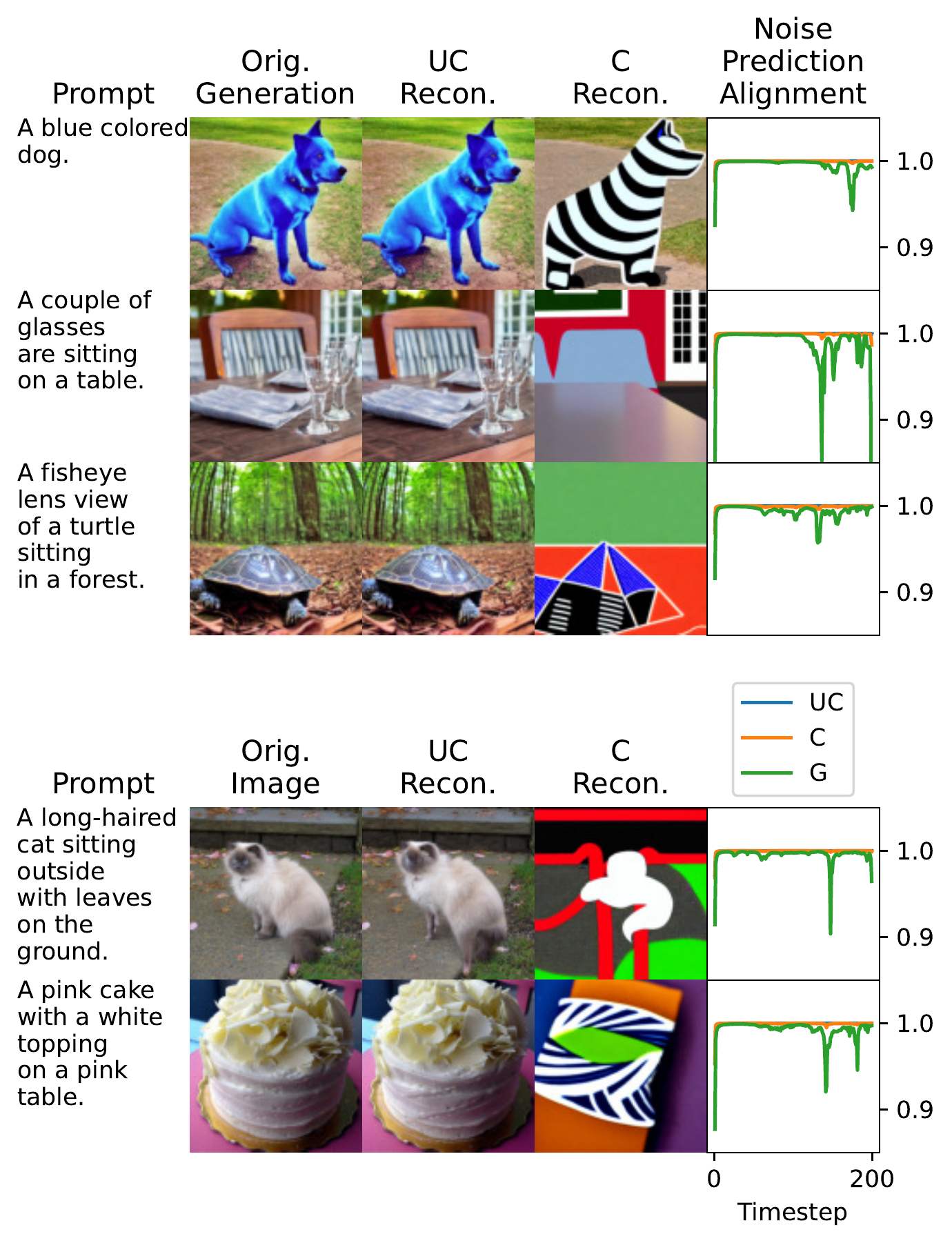}
    \caption{
    As \cref{fig:inversion_50} for 200 steps. 
    We observe that while the pseudo-gradient is more aligned between steps than previously, there are still large regions of misalignment (and given the number of steps these errors can accumulate).
    }
    \label{fig:inversion_200}
\end{figure*}

In Section 3.2 of the main paper we claim that the pseudo-gradient of classifier-free guidance $G \cdot (\Theta(x_t, t, C) - \Theta(x_t, t, \emptyset))$ is inconsistent across time steps which drives the instability of vanilla DDIM inversion and reconstruction results.
We demonstrate and analyze this instability in \cref{fig:inversion_50} (see caption).
We show that similar behavior holds for higher steps (\cref{fig:inversion_200}).

\section{Edits}\label{sec:sec3}

\subsection{Additional Edit Results}

\begin{figure*}
    \centering
    \includegraphics[width=\linewidth]{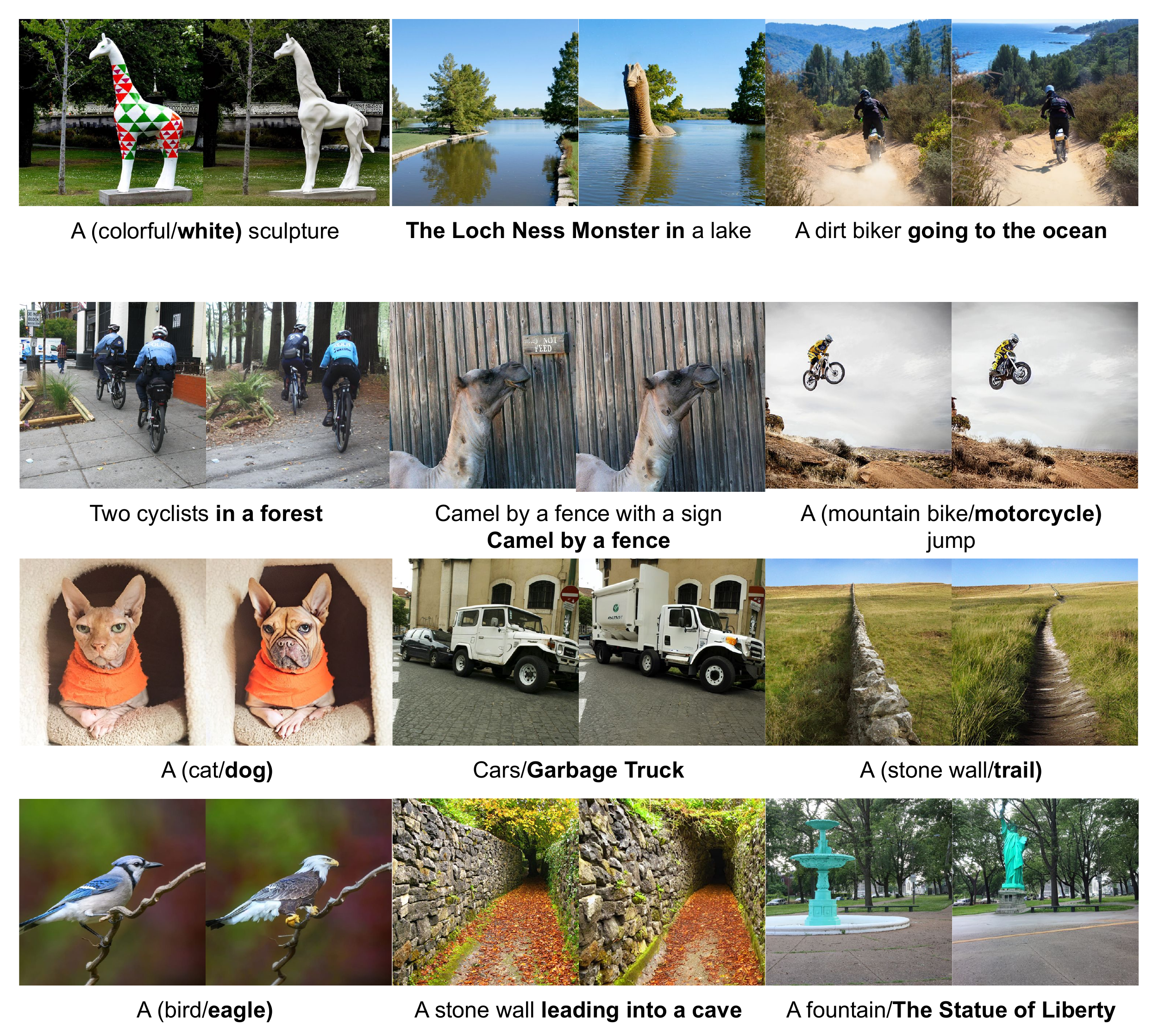}
    \caption{
    Additional edits demonstrating EDICT's versatility.
    \textbf{Bold} parts of prompt are the edit.
    }
    \label{fig:extra_edits}
\end{figure*}

\begin{figure*}
    \centering
    \includegraphics[width=\linewidth]{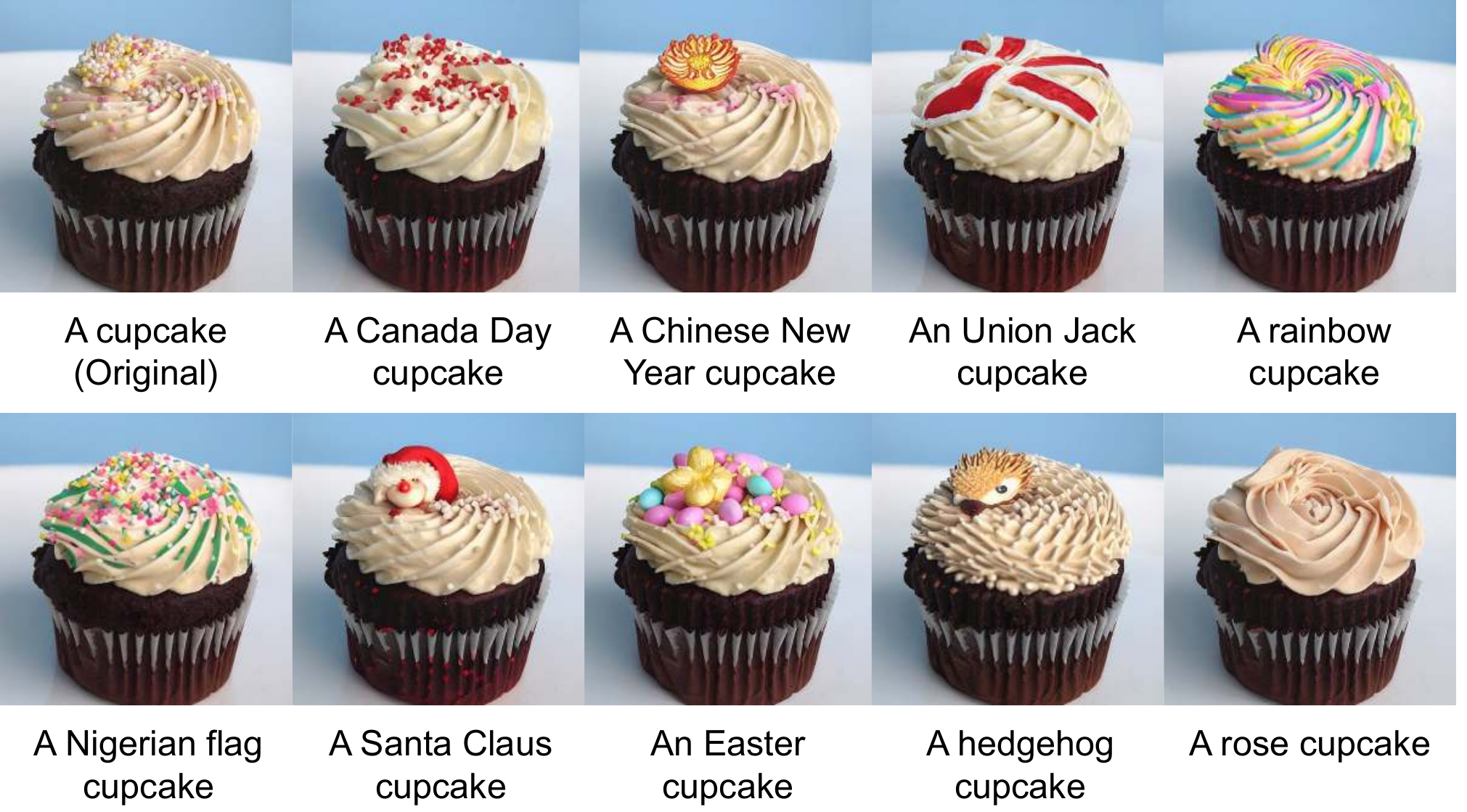}
    \caption{
    A variety of types of cupcake edited with EDICT.
    }
    \label{fig:cupcake_edits}
\end{figure*}

In \cref{fig:extra_edits} and \cref{fig:cupcake_edits} we display further editing results.

\subsection{Baselines with More Steps}

\begin{figure*}
    \centering
    \includegraphics[width=\linewidth]{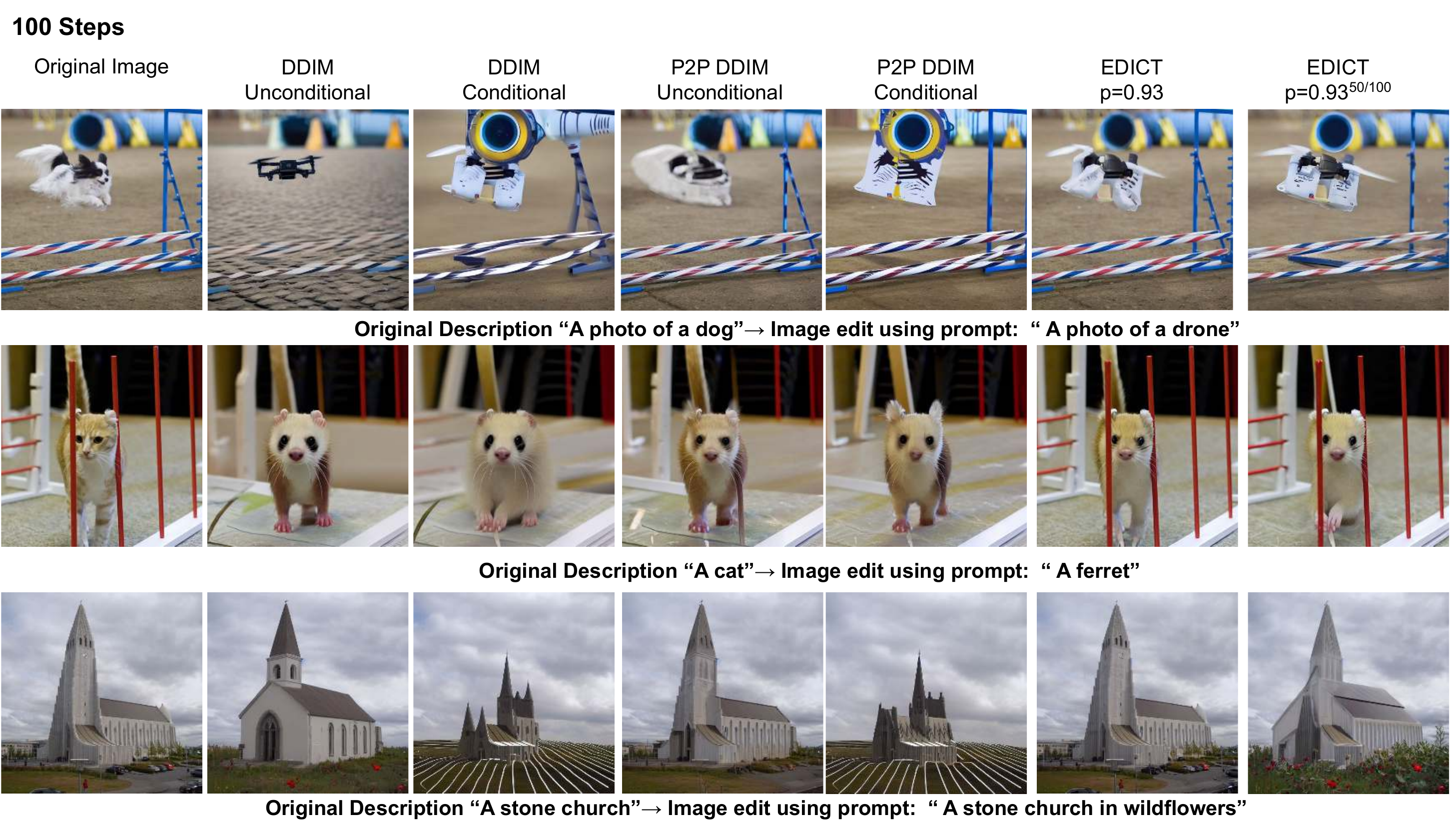}
    \caption{
    Baselines as in Figure 7 from the main body ran with steps $S=100$. We note that while some detail is lost for EDICT in the bottom row it still performs well compared to the other methods and maintains far superior performance in the top two rows.
    }
    \label{fig:baseline_comp100}
\end{figure*}

\begin{figure*}
    \centering
    \includegraphics[width=\linewidth]{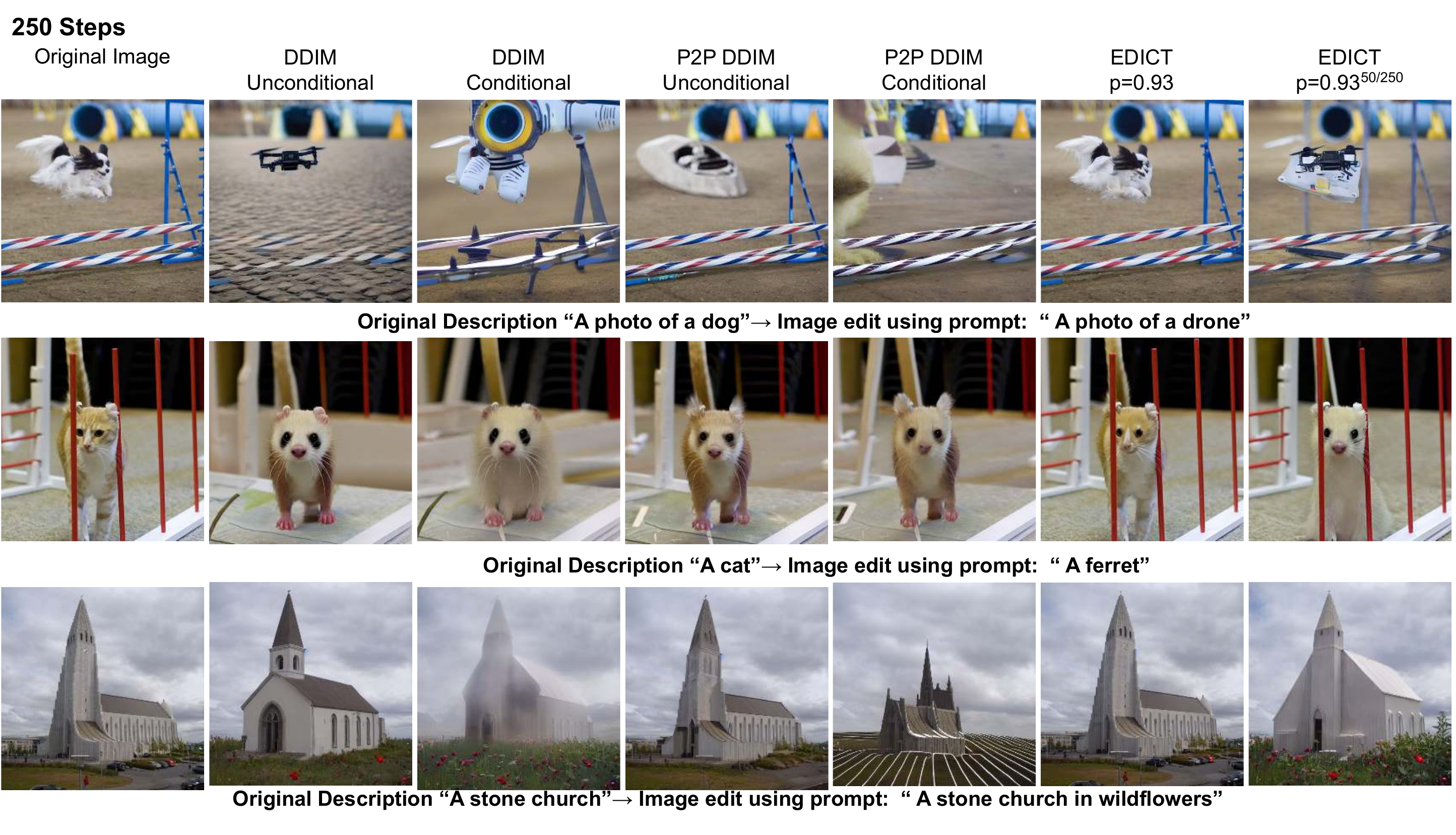}
    \caption{
    Baselines as in Figure 7 from the main body ran with steps $S=250$. We note that while some detail is lost for EDICT in the bottom row it still 
    }
    \label{fig:baseline_comp250}
\end{figure*}

In \cref{fig:baseline_comp100} and \cref{fig:baseline_comp250} we re-run the experiments of Figure 7 from the main paper with 100 and 250 global steps instead of the default 50.
We observe minimal changes besides some instability in the final row.
Note that we follow a scaling rule of $p=0.93^{50/S}$ to maintain the same aggregate dilation/contraction factor of $0.93^{50}$ from the original experiments.

\subsection{Dog Breeds: Extended Results}

\begin{figure*}
    \centering
    \includegraphics[width=\linewidth]{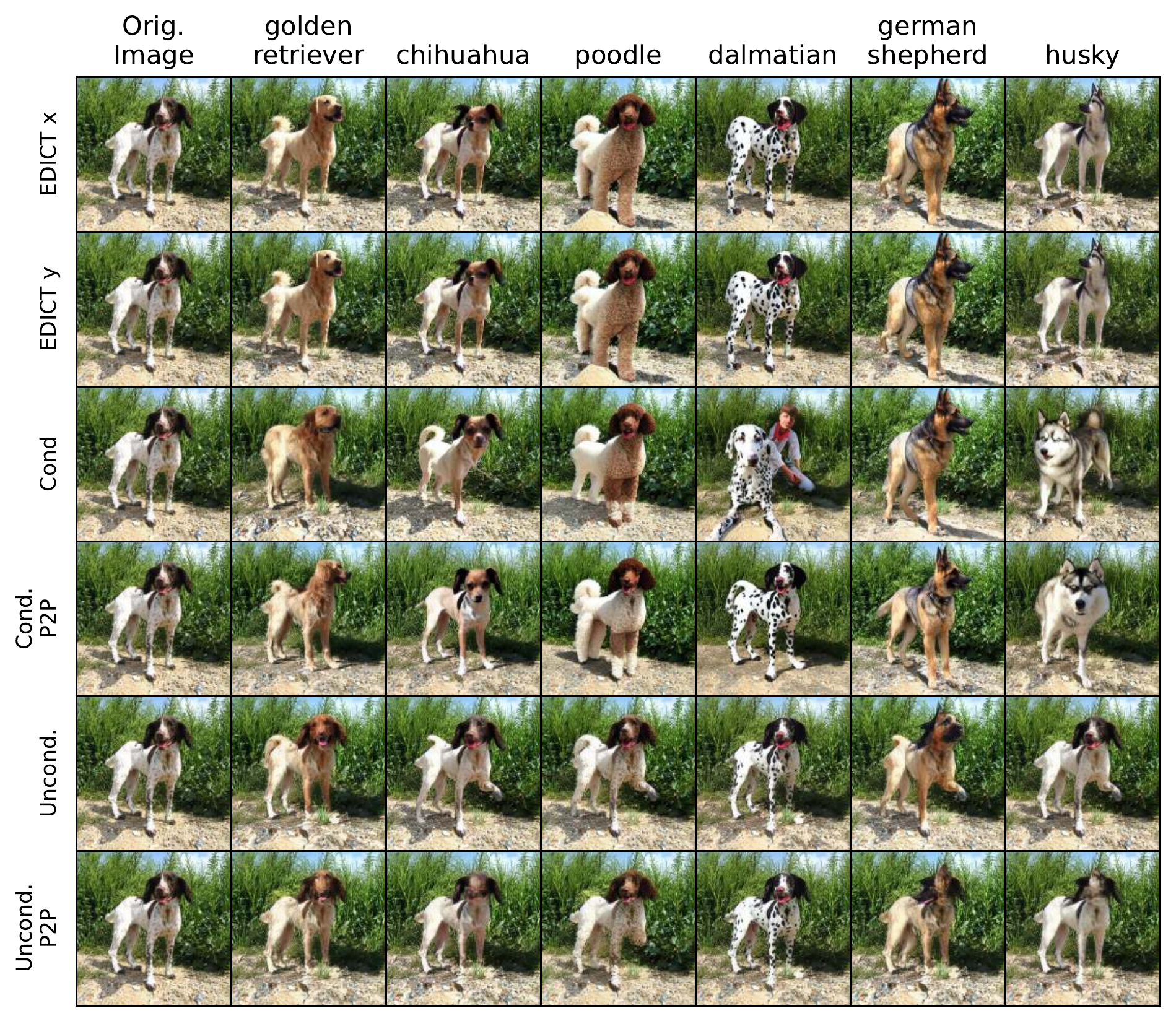}
    \caption{Dog breeds 1}
    \label{fig:dog1}
\end{figure*}

\begin{figure*}
    \centering
    \includegraphics[width=\linewidth]{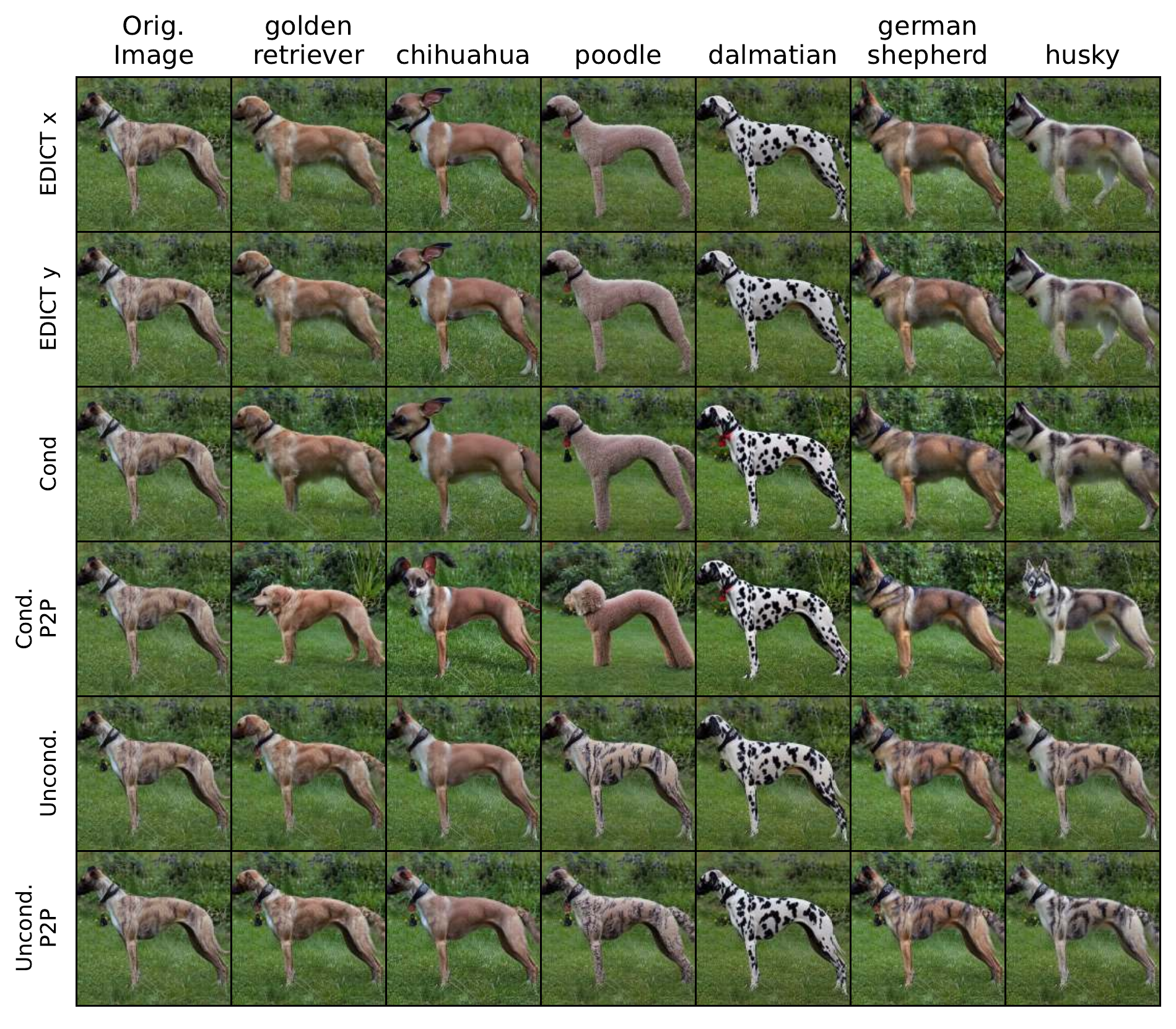}
    \caption{Dog breeds 2}
    \label{fig:dog2}
\end{figure*}

\begin{figure*}
    \centering
    \includegraphics[width=\linewidth]{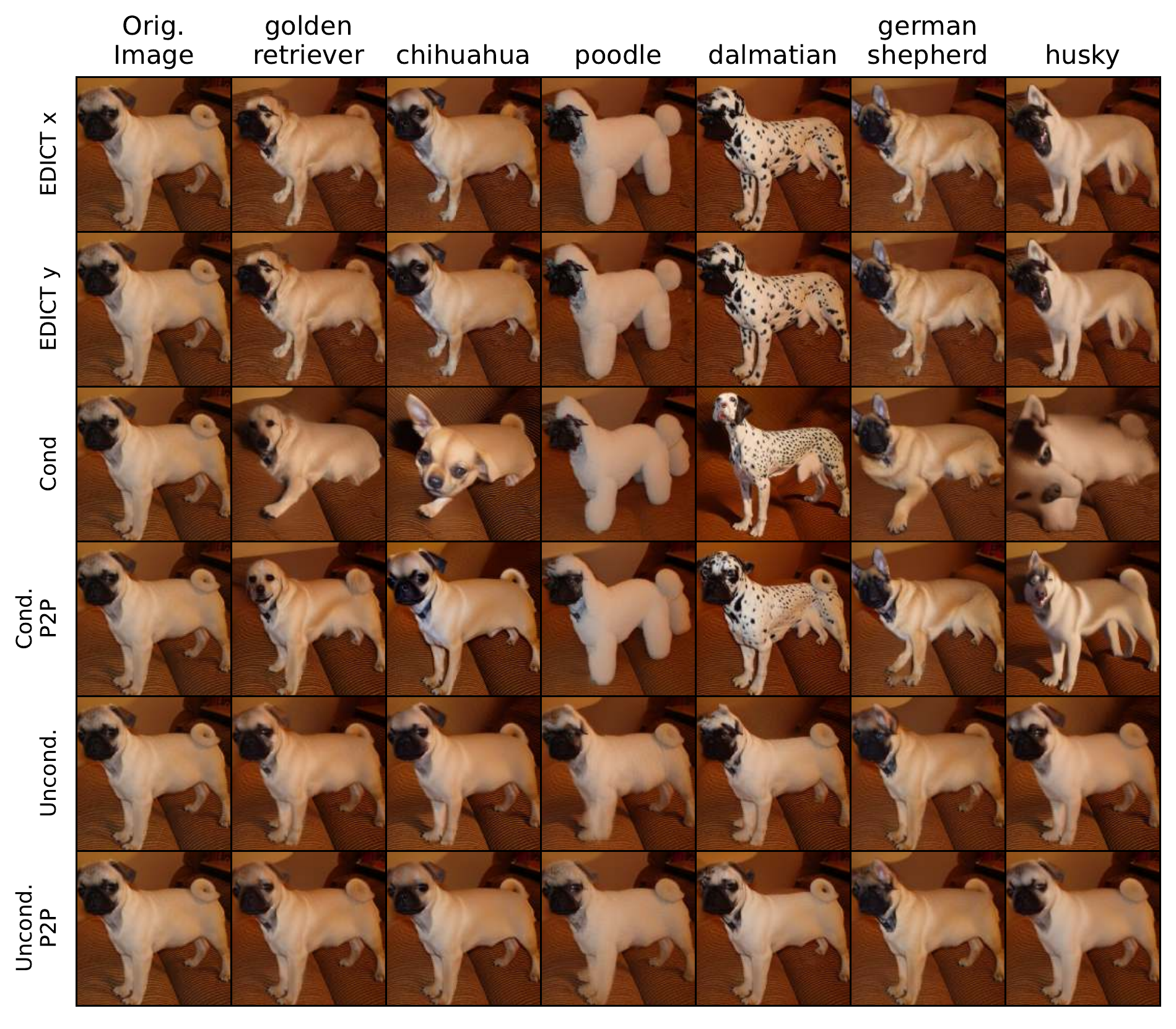}
    \caption{Dog breeds 3}
    \label{fig:dog3}
\end{figure*}

\begin{figure*}
    \centering
    \includegraphics[width=\linewidth]{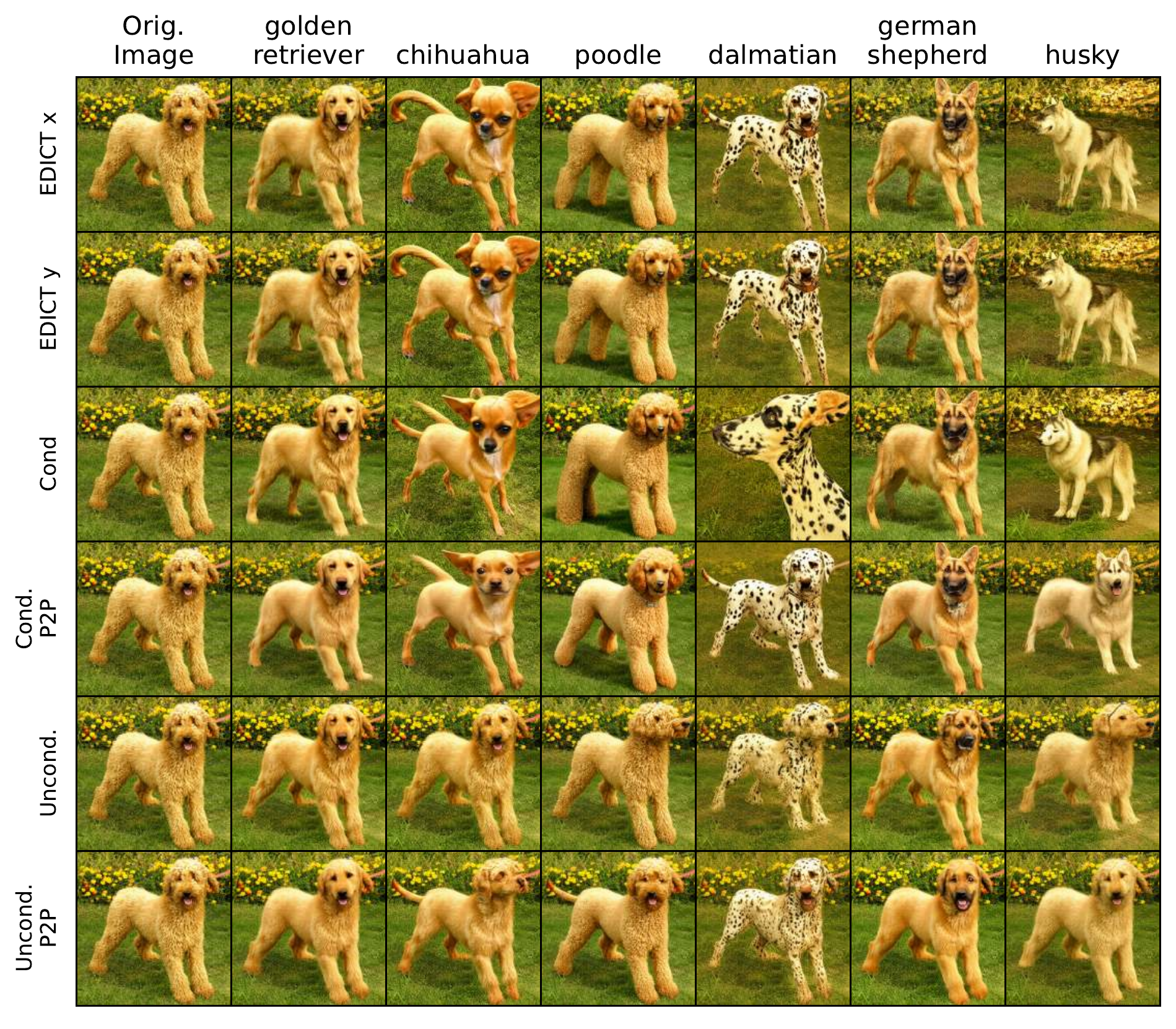}
    \caption{Dog breeds 4}
    \label{fig:dog4}
\end{figure*}

\begin{figure*}
    \centering
    \includegraphics[width=\linewidth]{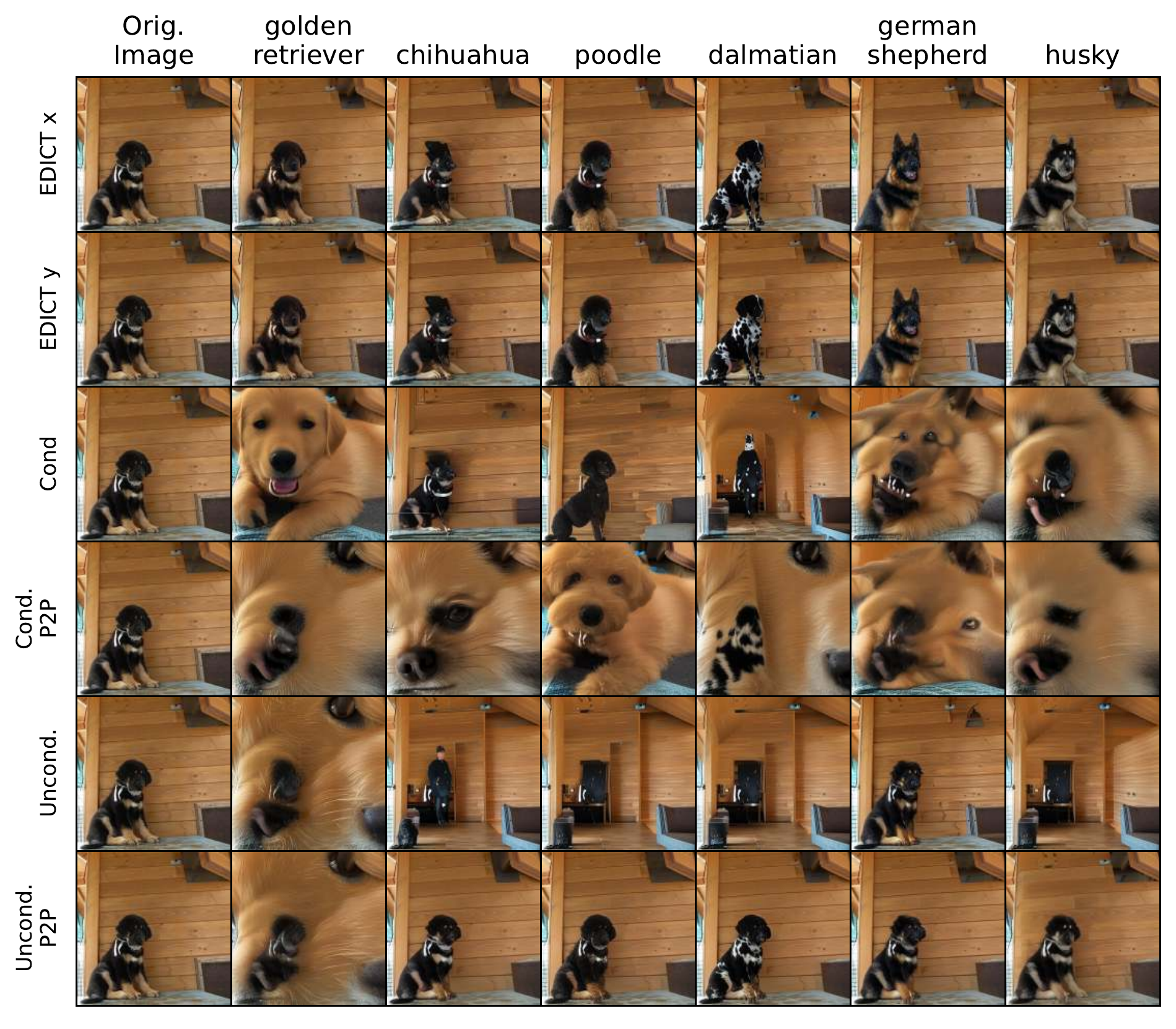}
    \caption{Dog breeds 5}
    \label{fig:dog5}
\end{figure*}

\begin{figure*}
    \centering
    \includegraphics[width=\linewidth]{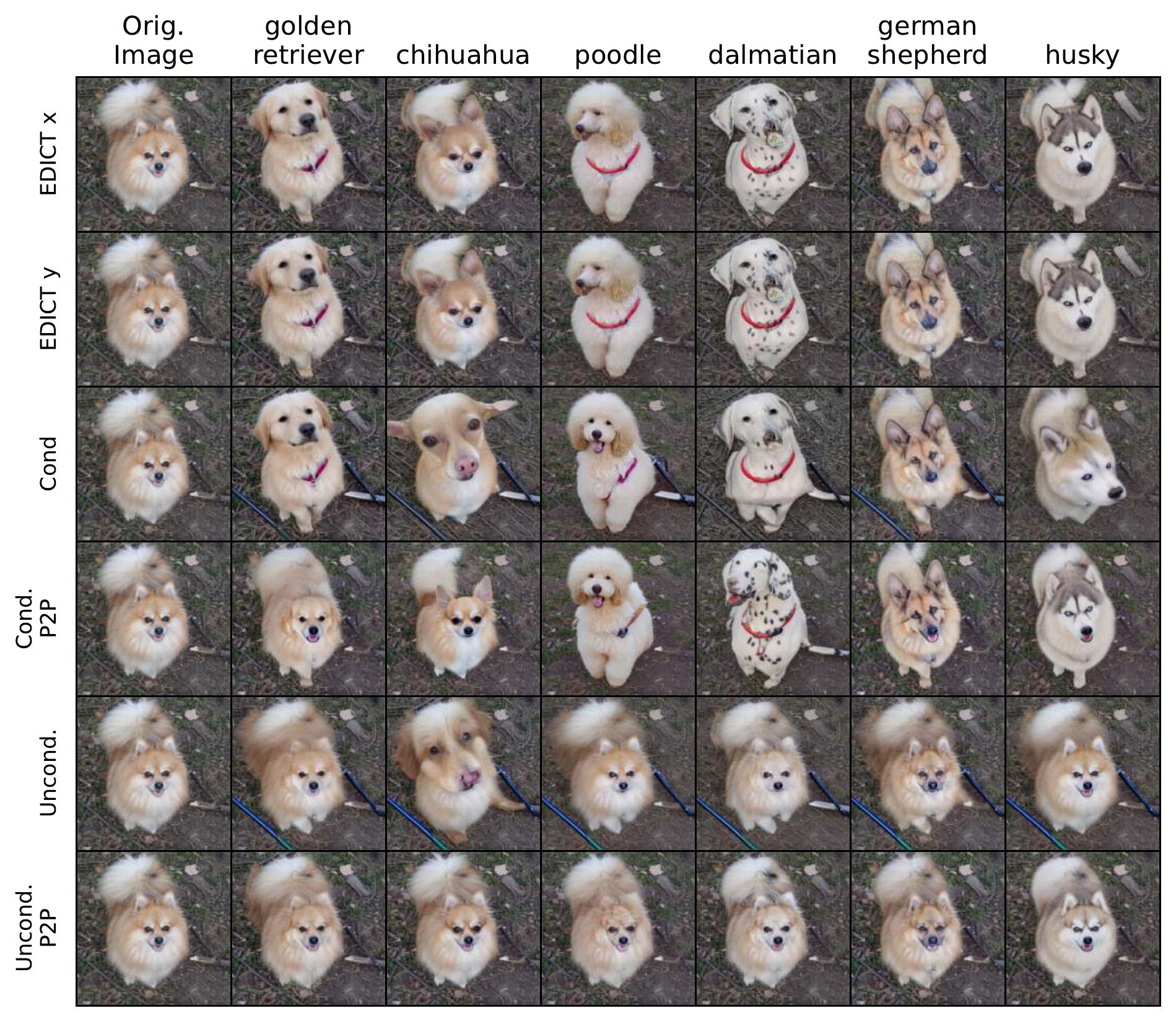}
    \caption{Dog breeds 6}
    \label{fig:dog6}
\end{figure*}

\begin{figure*}
    \centering
    \includegraphics[width=\linewidth]{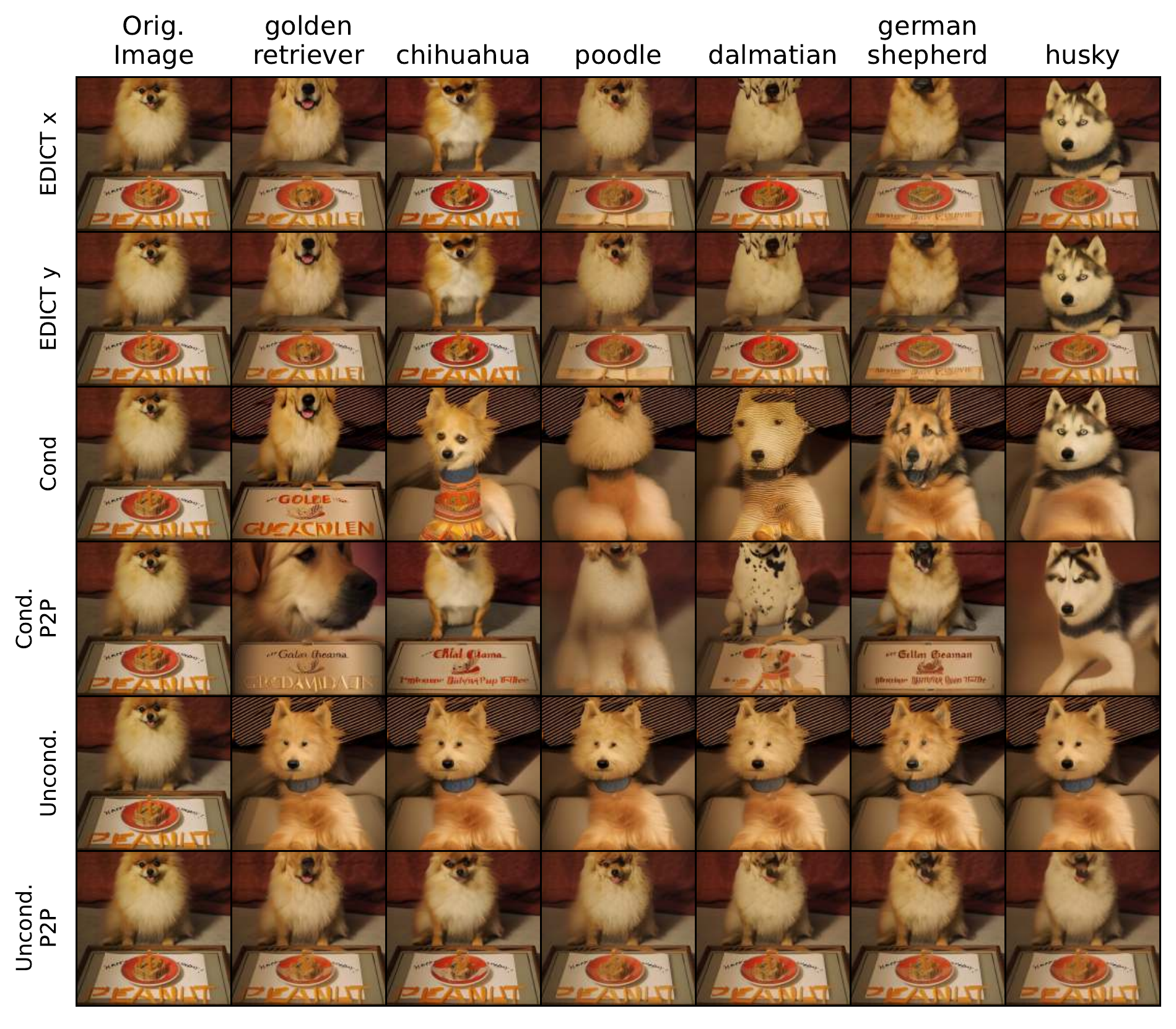}
    \caption{Dog breeds 7}
    \label{fig:dog7}
\end{figure*}

In \cref{fig:dog1}--\cref{fig:dog7} we display additional results of dog breed editing with baselines included. EDICT $x$ vs $y$ are the two sequence outputs of the EDICT process to demonstrate the visually-identical convergence.
We observe that EDICT consistently matches the desired output while preserving background details that baseline methods erase or alter. 
The base prompt is \textit{A dog} and the target prompt is \textit{A [target dog breed]}.


\section{Reconstruction Results}\label{sec:sec4}

In an extension of Table 1 from the main paper, we provide higher precision MSEs as well as reconstruction errors for 1000 steps in \cref{tab:recon_supp}.

 \begin{table*}
  \centering
  \resizebox{0.7\linewidth}{!}{
  \begin{tabular}{c|c c c c c}
  \multicolumn{6}{c}{COCO Reconstruction Error (MSE)} \\
    \toprule
    \multirow{2}{*}{Method} & \multirow{2}{*}{LDM AE} & EDICT & EDICT & DDIM & DDIM \\
    & & (UC) & (C) & (UC) & (C) \\
    \midrule
    50 Steps & 0.015260 & 0.015260 & 0.015260 & 0.030083 & 0.418175 \\
    200 Steps & 0.015260 & 0.015260 & 0.015260 & 0.023406 & 0.496944 \\
    1000 Steps & 0.015260 & 0.015260 & 0.015260 & 0.018960 & 0.509345 \\
    \bottomrule 
    
  \end{tabular} 
  }
  \caption{
  Mean-square error reconstruction results for the COCO validation set using the first listed prompt as conditioning with full-strength guidance. The latent diffusion model autoencoder (LDM AE), which is the autoencoder used to compute latents for reconstruction is the lower bound on reconstruction error.
  }
  \label{tab:recon_supp}
\end{table*} 

\end{document}